\documentclass[lettersize,journal]{IEEEtran}
\usepackage{amsmath,amsfonts}
\usepackage{algorithmic}
\usepackage{algorithm}
\usepackage{array}
\usepackage{textcomp}
\usepackage{stfloats}
\usepackage{url}
\usepackage{verbatim}
\usepackage{graphicx}
\usepackage{cite}
\usepackage[switch]{lineno}
\graphicspath{{./images/}}
\usepackage[pagebackref,breaklinks,colorlinks,citecolor=darkerblue]{hyperref}  

\usepackage{booktabs}       
\usepackage{nicefrac}       
\usepackage{microtype}      
\usepackage{listings}
\usepackage{float}
\usepackage{multirow}
\usepackage{subfigure}
\usepackage{wrapfig}
\usepackage{xcolor}  
\usepackage{colortbl}

\usepackage{mathtools} 
\usepackage{bbm} 
\usepackage{amsmath} 
\usepackage{chemformula}   
\usepackage{enumitem}  
\usepackage{amsfonts}   
\usepackage{caption}   
\usepackage{chemformula}   
\usepackage{algorithm,algorithmic}

\definecolor{darkergreen}{RGB}{21, 152, 56}
\definecolor{darkerblue}{rgb}{0,0.08,0.45}
\begin{document}
\title{Discovering the Representation Bottleneck of Graph Neural Networks}

\author{Fang Wu, Siyuan Li, Stan Z. Li 
\IEEEmembership{Fellow, IEEE} 
\thanks{Fang Wu is with the Computer Science Department, Stanford University, CA, USA. Siyuan Li and Stan Z. Li are with the School of Engineering, Westlake University, Zhejiang Province, China. E-mails: fangwu97@stanford.edu, lisiyuan@westlake.edu.cn, stan.zq.li@westlake.edu.cn}
\thanks{Fang Wu and Siyuan Li contributed equally to this work.}
\thanks{Corresponding Author: Stan Z. Li. \\
This work was supported by the project 2022ZD0115101. The authors acknowledge the financial support for covering the article processing charge.}
}
\markboth{IEEE Transactions on Knowledge and Data Engineering-2024}{}

\maketitle

\begin{abstract}
Graph neural networks (GNNs) rely mainly on the message-passing paradigm to propagate node features and build interactions, and different graph learning problems require different ranges of node interactions. In this work, we explore the capacity of GNNs to capture node interactions under contexts of different complexities. We discover that \emph{GNNs usually fail to capture the most informative kinds of interaction styles for diverse graph learning tasks}, and thus name this phenomenon GNNs' representation bottleneck.
As a response, we demonstrate that the inductive bias introduced by existing graph construction mechanisms can result in this representation bottleneck, \emph{i.e.}, preventing GNNs from learning interactions of the most appropriate complexity.  To address that limitation, we propose a novel graph rewiring approach based on interaction patterns learned by GNNs to dynamically adjust each node's receptive fields. Extensive experiments on both real-world and synthetic datasets prove the effectiveness of our algorithm in alleviating the representation bottleneck and its superiority in enhancing the performance of GNNs over state-of-the-art graph rewiring baselines.
\end{abstract}


\section{Introduction}
Graph neural networks (GNNs)~\cite{kipf2016semi,hamilton2017inductive} have witnessed growing popularity due to their ability to handle complex relationships and interdependence between objects, ranging from social networks~\cite{fan2019graph} to computer programs~\cite{nair2020funcgnn}. In particular, they show promise in scientific research and are used to derive insights from molecular structures~\cite{wu2018moleculenet} and reason about the relations in a group of interacting objects. Subsequently, efforts have been devoted to leveraging 3D geometry such as directions~\cite{klicpera2020fast} and dihedral angles~\cite{liu2021spherical}. 
Since physical rules remain stable regardless of the reference coordinate system, equivariance has been regarded as a ubiquitous property and integrated into GNNs with remarkable benefits~\cite{satorras2021n}.

GNNs' success provokes the succeeding bottleneck question: ``\emph{What are the common limitations of GNNs in real-world applications, such as molecules and dynamic systems?}'' Since the majority of GNNs are expressed as a neighborhood aggregation or message-passing scheme~\cite{gilmer2017neural,velickovic2017graph}, we leverage the tool of multi-order interactions between input variables~\cite{deng2021discovering} to investigate their representation bottleneck, aiming to analyze which types of interaction patterns (\emph{e.g.}, certain physical or chemical concepts) are likely to be encoded by GNNs, and which others are difficult to manipulate.

To this end, we formulate the metric of the multi-order interactions for GNNs from node-level and graph-level perspectives. With this computational instrument, we observe that the distribution of different kinds of node interactions learned by GNNs can deviate significantly from the initial data distribution of interactions. Furthermore, we relate GNNs' expressiveness with their capacity to capture interactions under different complexities. It is discovered that \emph{GNNs are typically incapable of learning the most informative interaction patterns and therefore cannot reach the global minimal loss point}, which we call the \textbf{representation bottleneck} of GNNs.

In this paper, we explain this representation bottleneck via graph construction mechanisms and observe that GNNs are more capable of encoding intermediate-complexity interactions. This discovery consists of the practice that pharmacologists are interested in identifying subgraphs (\emph{e.g.}, functional groups) that primarily represent specific molecular properties~\cite{yu2020graph,wang2021towards}.  Besides, we prove that existing methodologies to build node connectivity in scientific domains, such as \emph{K-nearest neighbor} (KNN) and \emph{fully connection} (FC) can introduce improper inductive bias. This improper inductive bias hidden inside the assumption of graph connectivity prohibits GNNs from encoding some particular interaction modes. 
To resolve the above-mentioned obstacle, we propose a novel graph rewiring technique based on the distribution of interaction strengths, named ISGR. First, it detects the interaction pattern that changes most violently, which is later used to progressively optimize the inductive bias of the GNNs and calibrate the topological structures of input graphs. Massive experimental evidence 
on synthetic and real-world datasets validate its considerable potential to ameliorate the representation bottleneck and achieve stronger interpretability and generalization for GNNs against all graph rewiring baselines.

Last, we revisit the representation behaviors of other categories of DNNs and compare them with GNNs. As a relevant answer, we observe the liability of CNNs to capture too simple pairwise interactions rather than more complex ones, which conforms to previous findings~\cite{deng2021discovering}. This inclination of CNNs can be explained by the inductive bias, as CNNs' fairly small kernel size assumes local connections between pixels or patches. However, as opposed to GNNs, CNNs are far more vulnerable to and seldom diverge from the original data distribution of node interactions. This analysis illustrates that both similarities and significant discrepancies exist between GNNs and other DNN classes in their representation activities. 
%

\section{Related Works}
\paragraph{GNNs' expressiveness and bottlenecks.} MLP is well-known for approximating any Borel measurable function~\cite{hornik1989multilayer}, but few study the universal approximation capability of GNNs.~\cite{hammer2005universal} demonstrates that cascade correlation can approximate functions with structured outputs.~\cite{scarselli2008computational} prove that a RecGNN~\cite{scarselli2008graph} can approximate any function that preserves unfolding equivalence to any degree of precision.~\cite{maron2018invariant} show that an invariant GNN can approximate an arbitrary invariant function defined on graphs. ~\cite{xu2018powerful} show that common GNNs are incapable of differentiating different graph structures. They further prove that if the aggregation and readout functions of a GNN are injective, it is at most as powerful as the Weisfieler-Lehman (WL) test~\cite{leman1968reduction} in distinguishing different graphs. 

Despite their potential in modeling structural data, GNNs capture only a tiny fragment of first-order logic~\cite{barcelo2020logical}, which arises from the deficiency of the receptive field of a node. Meanwhile, GNNs do not benefit from the increase of layers due to \emph{over-smoothing}~\cite{li2018deeper,klicpera2018predict,chen2020measuring} and \emph{over-squashing}~\cite{alon2020bottleneck,topping2021understanding}. Several works have attempted to overcome these issues by topology-imbalance optmization~\cite{sun2022position}, mixup~\cite{lu2024nodemixup}, weight normalization~\cite{oono2019graph}, and node skipping~\cite{lu2024skipnode}. To the best of our knowledge, we are the first to understand GNNs' expressiveness from interactions under different contextual complexities and link the expressive limitation with the inductive bias of graph connectivity.  

\paragraph{GNNs' representation capacity.} It becomes an emerging area to evaluate the representation capability of DNNs~\cite{shwartz2017opening,neyshabur2017exploring,novak2018sensitivity,weng2018evaluating,fort2019stiffness}. Interactions between variables are pioneeringly used to inspect the limitation of DNNs in feature representations~\cite{zhang2020interpreting,deng2021discovering,lundberg2017unified}. Previous researches mainly study the theoretically maximum complexity~\cite{shwartz2017opening}, generalization ability~\cite{novak2018sensitivity,weng2018evaluating,fort2019stiffness}, and robustness~\cite{neyshabur2017exploring} of DNNs.
Notwithstanding, prior works merely highlight the behaviors of general DNNs and examine their assertions via MLP and CNNs. In comparison, we emphasize GNNs that operate on structured graphs, distinct from images and texts.

\paragraph{Graph rewiring.} Rewiring is a process of altering the graph structure to control the information flow. Among existing approaches such as connectivity diffusion~\cite{klicpera2019diffusion}, bridge-node insertion~\cite{battaglia2018relational}, positional encoding~\cite{bruel2022rewiring}, and multi-hop filters~\cite{frasca2020sign}, edge sampling shows great power in tackling \emph{over-smoothing} and \emph{over-squashing}. The sampling strategies can be random drop~\cite{huang2020tackling} or based on edge relevance~\cite{klicpera2019diffusion,kazi2022differentiable}. For example,~\cite{alon2020bottleneck} modified the last layer to an FC-graph to help GNNs grasp long-range interactions. Moving further,~\cite{topping2021understanding} proves that negatively curved edges are responsible for \emph{over-squashing} and introduces a curvature-based rewiring method to alleviate that. Differently, our rewiring algorithm originates from a completely new motivation, \emph{i.e.}, reshaping graph structure to assist GNNs in learning the most informative order of interactions.

\section{Multi-order Interactions for Graphs}
\paragraph{Preliminary for GNNs}
Suppose a graph $\mathcal{G}=(\mathcal{V}, \mathcal{E})$ has a set of $n$ variables (\emph{a.k.a.} nodes). $\mathcal{G}$ can be a macroscopic physical system with $n$ celestial bodies, or a microscopic biochemical system with $n$ atoms, denoted as $\mathcal{V}=\{v_1,...,v_n\}$. 
$f$ is a well-trained GNN model and $f(\mathcal{G})$ represents the model output. For node-level tasks, the GNN forecasts a value (\emph{e.g.}, atomic energy) or a vector (\emph{e.g.}, atomic force or velocity) for each node. For graph-level tasks, $f(\mathcal{G})\in \mathbb{R}$ is a scalar (\emph{e.g.}, drug toxicity or binding affinity). Most GNNs make predictions by interactions between input nodes instead of working individually on each vertex. Accordingly, we concentrate on pairwise interactions and use the multi-order interaction $I^{(m)}(i,j)$~\cite{tsang2017detecting} to measure interactions of different complexities between two nodes $v_i,v_j\in \mathcal{V}$. 

\paragraph{Graph-level Multi-order Interactions} 
Specifically, the $m$-th order interaction $I^{(m)}(i,j)$ measures the average interaction utility between nodes $v_i$ and $v_j$ under all possible subgraphs $\mathcal{G}_S$, which consists of $m$ nodes. Mathematically, the multi-order interaction is defined as:
    \begin{equation}
    \label{equ: multi_order_interaction}
        I^{(m)}(i, j)=\mathbb{E}_{\mathcal{G}_S \subseteq \mathcal{G},\{v_i,v_j\}\subseteq \mathcal{V}, \mid \mathcal{V}_S \mid=m}[\Delta f(v_i, v_j, \mathcal{G}_S)],
    \end{equation}
where $3\leq m\leq n$ and $\Delta f(v_i, v_j, \mathcal{G}_S)$ is defined as $f(\mathcal{G}_S)-f(\mathcal{G}_S \backslash v_i)-f(\mathcal{G}_S \backslash v_j)+f(\mathcal{G}_S\backslash \{v_i, v_j\})$. $\mathcal{G}_S\subset \mathcal{G}$ is the context subgraph. $f(\mathcal{G}_S)$ is the GNN output when we keep nodes in $\mathcal{G}_S$ unchanged but delete others in $\mathcal{G}\backslash \mathcal{G}_S$. Since it is irrational to feed an empty graph into a GNN, we demand the context $S$ to have at least one variable with $m\geq 3$ and omit $f(\emptyset)$. Note that some studies~\cite{zhang2020interpreting} assume the target variables $v_i$ and $v_j$ do not belong to the context $\mathcal{G}_S$. Contrarily, we propose to interpret $m$ as the contextual complexity of the interaction and include nodes $v_i$ and $v_j$ in the subgraph $\mathcal{G}_S$. The proof is provided in the Appendix that the two cases are equivalent but from different views. An elaborate introduction of $I^{(m)}$ (\emph{e.g.}, the connection with existing metrics) is in the Appendix.

\paragraph{Node-level Multi-order Interaction}
$I^{(m)}(i,j)$ in Equ.~\ref{equ: multi_order_interaction} is designed to analyze the influence of interactions over the integral system (\emph{e.g.}, a molecule or a galaxy) and is therefore only suitable in the circumstance of graph-level prediction. However, many graph learning problems are node-level tasks. For the sake of measuring the effects of those interactions on each component (\emph{e.g.}, atom or particle) of the system, we introduce a new metric as follows:
\begin{equation}
\label{equ: interaction_node}
    I^{(m)}_i(j)=\mathbb{E}_{\mathcal{G}_S \subseteq \mathcal{G}, v_j\in \mathcal{V},\mid\mathcal{V}_S\mid=m}[\Delta f_i(v_j, \mathcal{G}_S)],
\end{equation}
where $2\leq m\leq N$. $\Delta f_i(v_j, \mathcal{G}_S)$ is formulated as $\left\|f_i(\mathcal{G}_S)-f_i(\mathcal{G}_S \backslash v_j)\right\|_p$, and $\|.\|_p$ is the $p$-norm. We denote $f_i(\mathcal{G}_S)$ as the output for node $v_i$ when other nodes in $\mathcal{G}_S$ are kept unchanged. Equ.~\ref{equ: interaction_node} allows us to measure the representation capability of GNNs in node-level tasks.

\paragraph{Interaction Strengths} 
To measure the reasoning complexity of GNNs, we compute the graph-level relative interaction strength $J^{(m)}$ of the encoded $m$-th order interaction as follows:
\begin{equation}
\label{equ: strength}
    J^{(m)}=\frac{\mathbb{E}_{\mathcal{G} \in \Omega}\left[\mathbb{E}_{i, j}\left[\mid I^{(m)}(i, j \mid \mathcal{G})\mid\right]\right]}{\sum_{m^{\prime}}\left[\mathbb{E}_{\mathcal{G} \in \Omega}\left[\mathbb{E}_{i, j}\left[\mid I^{\left(m^{\prime}\right)}(i, j \mid \mathcal{G})\mid \right]\right]\right]},
\end{equation}
where $\Omega$ stands for the set of all graph samples, and the strength $J^{(m)}$ is calculated over all pairs of input variables in all data points. {Remarkably, the distribution of $J^{(m)}$ measures the distribution of the complexity of interactions encoded in GNNs.} Then we normalize $J^{(m)}$ by the summation value of $I^{(m)}(i, j \mid \mathcal{G})$ with different orders to constrain $0\leq J^{(m)}\leq 1$ for explicit comparison across various tasks and datasets. 
Then the node-level interaction strength is defined as $J^{(m)}=\frac{\mathbb{E}_{\mathcal{G} \in \Omega}\left[\mathbb{E}_{i}\left[\mathbb{E}_{j}\left[\mid I^{(m)}_i(j \mid \mathcal{G})\mid \right]\right]\right]}{\sum_{m^{\prime}}\left[\mathbb{E}_{\mathcal{G} \in \Omega}\left[\mathbb{E}_{i}\left[\mathbb{E}_{j}\left[\mid I^{\left(m^{\prime}\right)}_i(j \mid \mathcal{G})\mid\right]\right]\right]\right]}$.

According to the efficiency property of $I^{(m)}(i, j)$~\cite{deng2021discovering}, the change of GNN parameters $\Delta W$ can be decomposed as the sum of gradients ${\partial I^{(m)}(i, j)}/{\partial W}$. Mathematically, we denote $L$ and $\eta$ as the loss function and learning rate. With $U=\sum_{v_i \in \mathcal{V}} f(v_i)$ and $R^{(m)}=-\eta \frac{\partial L}{\partial f(\mathcal{G})} \frac{\partial f(\mathcal{G})}{\partial I^{(m)}(i, j)}$, it is attained by: 
the output of GNNs can be explained as the sum of all interaction utilities as $f(\mathcal{G})=\sum_{i \in N} f(\{i\})+\sum_{i, j \in N, i \neq j} \sum_{m=3}^{n} \frac{n+1-m}{n(n-1)} I^{(m)}(i, j)$. 
\begin{equation}
\label{equ: delta_W}
\begin{split}
    \Delta W& =-\eta \frac{\partial L}{\partial W}= -\eta \frac{\partial L}{\partial f(\mathcal{G})} \frac{\partial f(\mathcal{G})}{\partial W} \\
    & =\Delta W_{U}+\sum_{m=3}^{n} \sum_{v_i, v_j \in \mathcal{V}, v_i \neq v_j} R^{(m)} \frac{\partial I^{(m)}(i, j)}{\partial W}.
\end{split}
\end{equation}

\begin{figure*}[t]
\centering
\subfigure[QM7]{\includegraphics[width=2.25in]{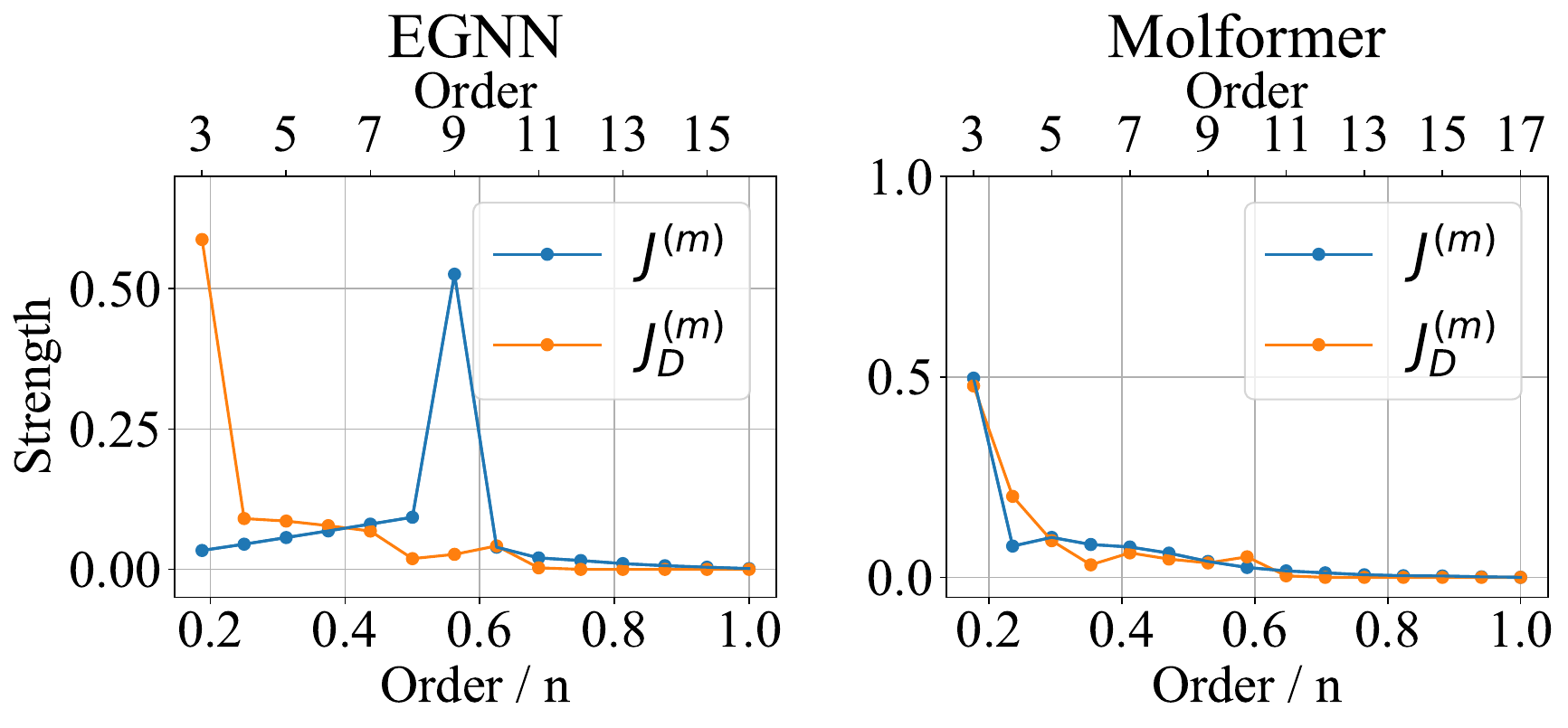} } 
\subfigure[QM8]{\includegraphics[width=2.25in]{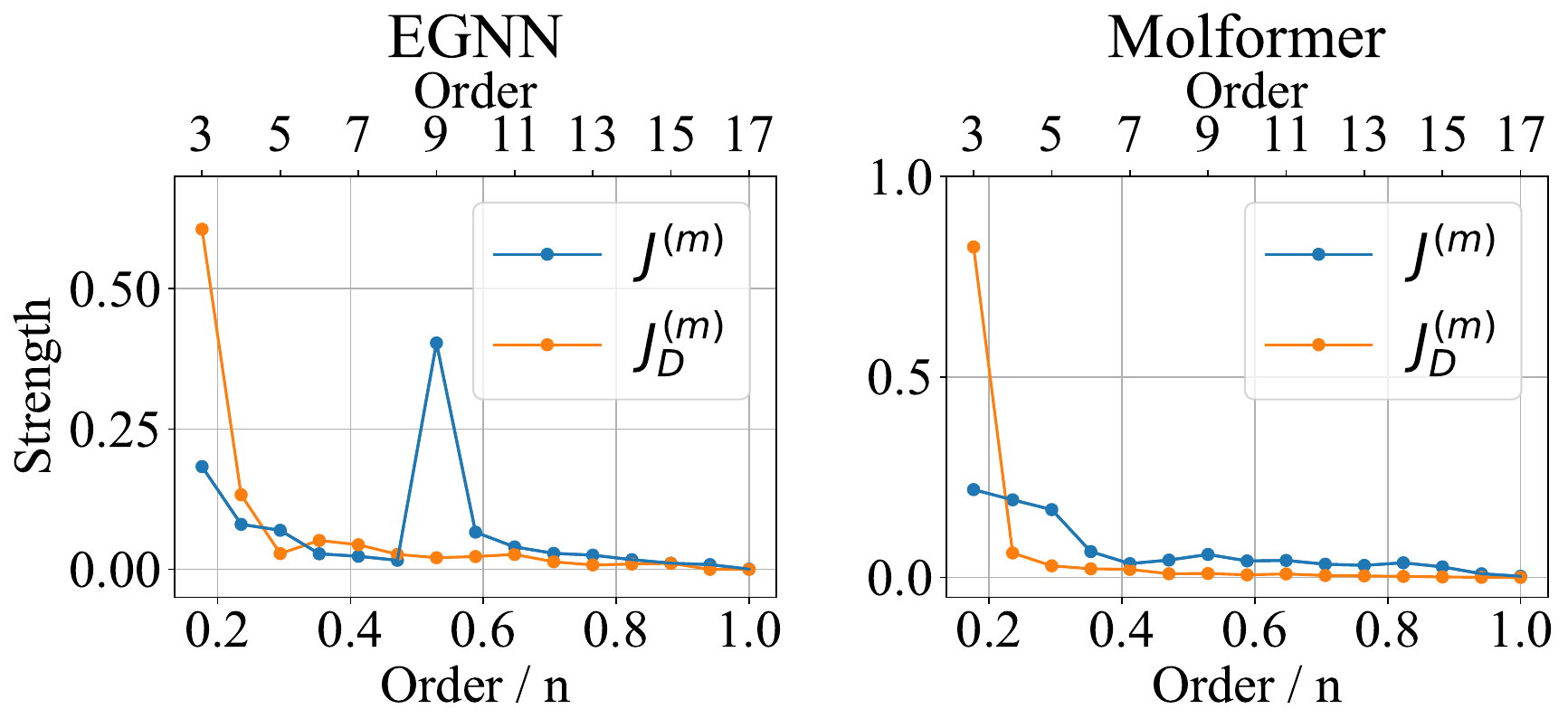} } 
\subfigure[Hamiltonian Dynamics]{\includegraphics[width=2.25in]{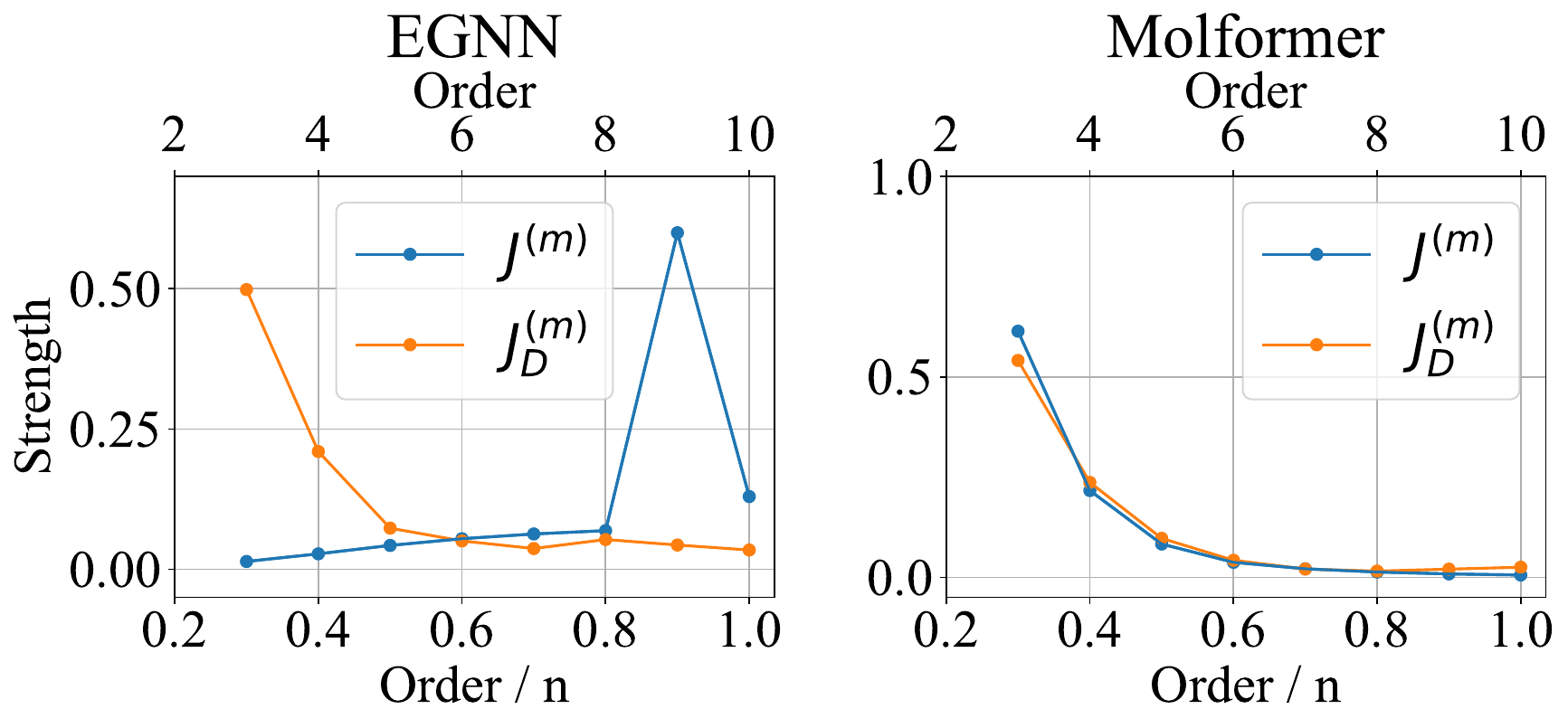} }
\subfigure[Newtonian Dynamics]{\includegraphics[width=2.25in]{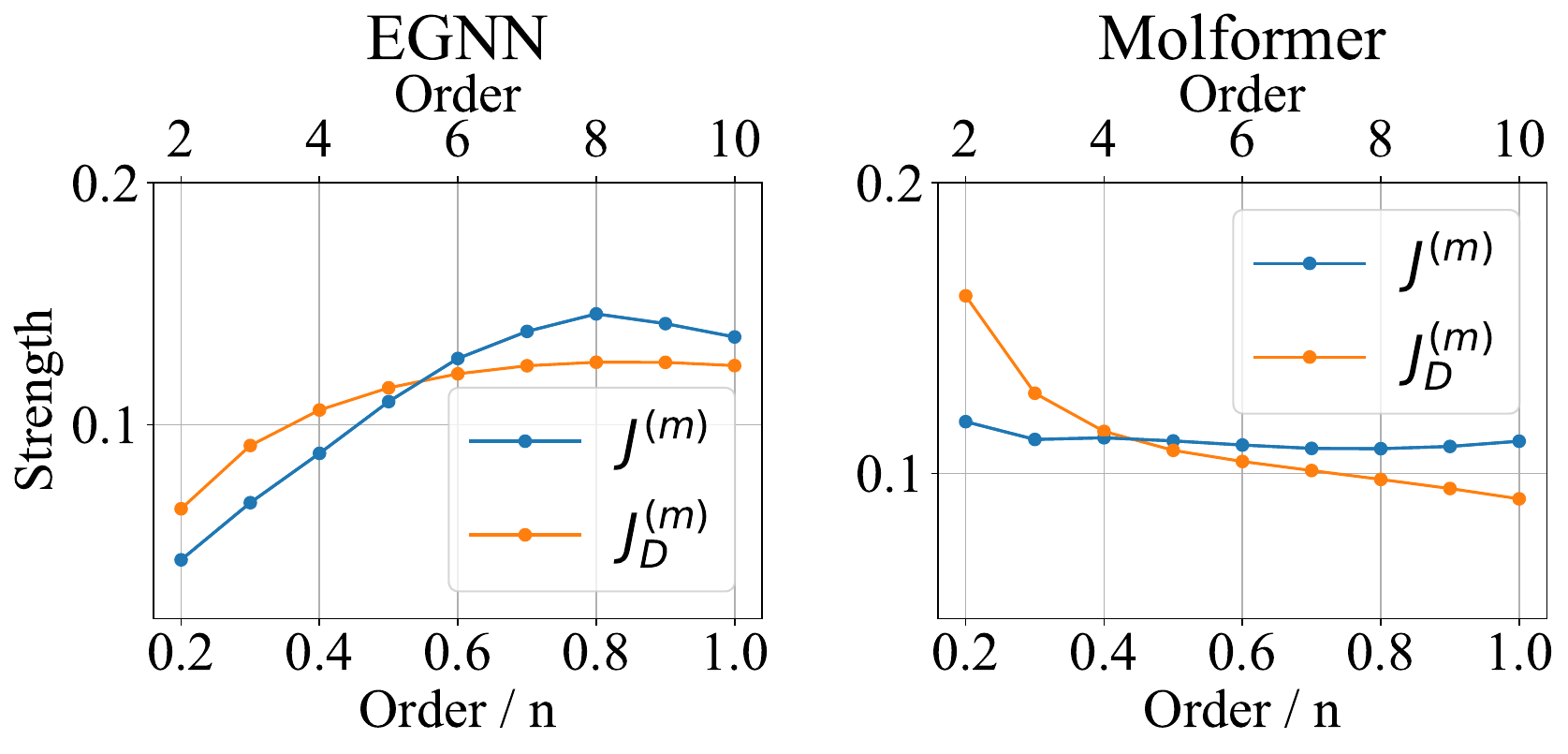}}
\subfigure[Molecular Dynamics]{\includegraphics[width=2.25in]{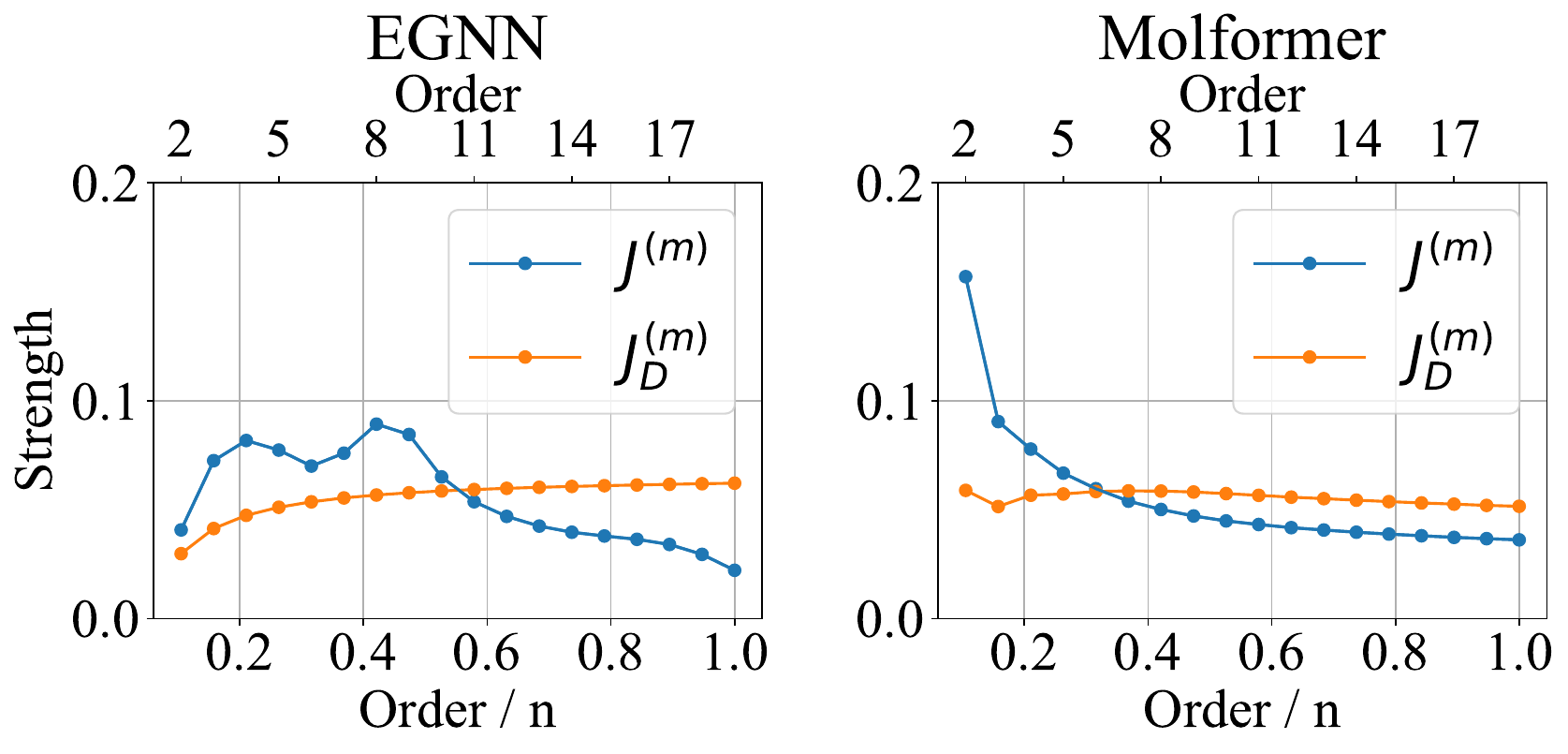}}
\caption{Distributions of interaction strengths of EGNN and Molformer in graph-level and node-level prediction tasks. We use double-x axes to represent the order $m$ (bottom) and the ratio $m/n$ (top).}
\label{fig: str_order}
\end{figure*}

\section{Representation Bottleneck of GNNs}
\label{sec: rep_gnns}
\subsection{Definition of Representation Bottleneck}
For modern GNNs, the loss $L$ is typically non-convex with multiple local and even global minima~\cite{foret2020sharpness}. This diversity of minima may yield similar values of $L$ while acquiring different capacities of GNNs to learn interactions. That is, each loss point must correspond to a different $J^{(m)}$. Intuitively, we declare that if $J^{(m)}$ learned by a GNN model $f$ is not equivalent to the optimal strength ${J^{(m)}}^*$, then $f$ must be stuck in a local minimum point of the loss surface.
In particular, different tasks require diverse ranges of interactions. For instance, short-range interactions play a more critical role in small-molecule binding prediction~\cite{wu2022pre}, while protein-protein rigid-body docking attaches more significance to the middle-range or long-range interactions between residues. Thus, ${J^{(m)}}^*$ varies according to particular tasks and datasets. 
Based on the relationship between $J^{(m)}$ and the training loss $l$, we define the representation bottleneck of GNNs as the phenomenon that \emph{the distribution of interaction strength $J^{(m)}$ learned by GNNs fails to reach the optimal distribution of the interaction strength ${J^{(m)}}^*$}. 

\subsection{The Role of Inductive Bias}
Many factors can instigate this GNN's representation bottleneck. In this paper, we focus on the improper inductive bias introduced by the graph construction mechanism as a partial answer. 

\paragraph{GNNs for Scientific Problems}
GNNs operate on graph-structured data and have strong links to the field of geometric deep learning. Aside from studies on social networks~\cite{fan2019graph} and citation networks~\cite{tang2008arnetminer} as well as knowledge graphs~\cite{carlson2010toward}, science, including biology, physics, and chemistr,y has been one of the main drivers in the development of GNNs~\cite{mrowca2018flexible,sanchez2020learning,wu2022geometric}. In this work, our focus is on analyzing the representation behavior of geometric GNNs for scientific explorations.  

Notably, graphs in most scientific problems are unlike common applications of GNNs such as recommendation systems~\cite{wu2022graph} and relation extraction~\cite{zhu2019graph}. Indeed, molecules or dynamic systems do not have explicit edges. 
\emph{KNN}, \emph{full connections} (FC), and \emph{r-ball} are the three most broadly used mechanisms to build node connectivity. 
KNN-graphs build edges based on pairwise distances in the 3D space and are a common technique in the modeling of macromolecules~\cite{ganea2021independent,stark2022equibind}.
FC-graphs, instead, assume all nodes are connected to each other~\cite{chen2019path,baek2021accurate, AlphaFold2021}. 
In \emph{r-ball} graphs, an edge exists between any node pair as long as its spatial distance is shorter than a threshold value. 

\begin{figure}[ht]
    \centering
    \includegraphics[scale=0.60]{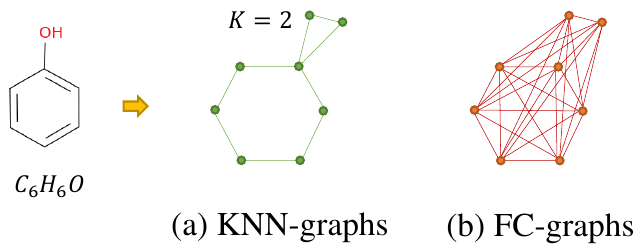}
    \caption{Different graph constructions of the compound $\ch{C_6H_6O}$.}
    \label{fig:graph_types}
\end{figure}

To align with the multi-order interaction theory, graph construction must satisfy two properties: (1) The subgraph maintains connectivity. (2) No ambiguity is introduced from either the structural or feature view. However, \emph{r}-ball graphs do not satisfy the first property. If node $a$ is merely linked to node $b$, it would be isolated if $b$ is removed from the graph. In contrast, subgraphs in KNN-graphs can be reconstructed via KNN to ensure connectivity, while the removal of any node in FC-graphs will not influence the association of other entity pairs. Hence, we only consider KNN- and FC-graphs in the subsequent analysis.
Here, we provide an example for constructing molecules or physical systems as graphs with three discussed methods. As shown in Fig.~\ref{fig:graph_types}, FC-graphs are a special type of KNN-graphs, where $K\geq n-1$. But FC-graphs or KNN-graphs with a large $K$ suffer from high computational expenditure and are usually infeasible with thousands of entities. Meanwhile, they are sometimes unnecessary since the impact from distant nodes is so minute to be ignored.

\paragraph{Inductive Bias of GNNs}
We define the data distribution of interaction strengths in a dataset $D$, denoted as $J^{(m)}_D$, as the experimental distribution of interaction strengths for model $f$ with randomly initialized parameters, which is later used for comparison between GNNs and other DNNs. Fig.~\ref{fig: str_order} reports the learned distributions $J^{(m)}$ and the data distributions $J^{(m)}_D$ for both graph-level and node-level tasks, where EGNN~\cite{satorras2021n} works on KNN-graphs and Molformer~\cite{wu20213d} performs on FC-graphs. The complete experimental details are elaborated in subsection~\ref{sec: exp_detail}.  

Based on these empirical plots, it can be observed that EGNN and Molformer are inclined to learn interactions of distinct orders. To be specific, EGNN is more prone to pay attention to interactions of some particular orders, and its $J^{(m)}$ lines usually have one or two spikes. On the contrary, Molformer learns a more unconstrained $J^{(m)}$. Its $J^{(m)}$ on Newtonian dynamics is extremely smooth, like a straight line, but its $J^{(m)}$ on Hamiltonian and molecular dynamics (MD) are steep curves. 

We owe this phenomenon to different inductive biases brought by different graph construction approaches. Namely, the inductive bias brought by the topological structure of input graphs significantly impacts $J^{(m)}$ of GNNs. Noticeably, we demonstrate in Sec.~\ref{sec: revisit} that CNNs can be forbidden from capturing appropriate orders of interactions due to their inductive bias of locality. This local characteristic comes from CNNs' relatively small kernel size. 
As for GNNs, their representation manners can also be influenced by the inductive bias, which primarily depends on graph connectivity. Unequivocally, FC-graphs consist of all pairwise relations and hypothesize that all particles can affect each other directly. This hypothesis places weaker restrictions on Molformer's representation behavior and results in more diverse $J^{(m)}$. On the contrary, KNN-graphs assume that some pairs of entities possess a connection and others do not. In our setting, the number of nearest neighbors $K$ for KNN-graphs is kept as $K=8$, which is equivalent to the order that has the highest strength in $J^{(m)}$ of EGNN.

Undeniably, inductive biases introduced by graph construction mechanisms can be improper and give rise to poor $J^{(m)}$ that is far away from ${J^{(m)}}^*$ after training. Prominently, bad inductive bias can impose a much larger impact on GNNs on $J^{(m)}$ than on CNNs. This is because graphs support arbitrary pairwise relational structures~\cite{battaglia2018relational}, and accordingly, the inductive bias of GNNs is more flexible and influential. In particular, KNN-graphs are more susceptible to improper inductive bias. EGNN hardly learns interactions of orders different from the user-defined constant $K$, which can cause worse performance. However, FC-graphs 
are not a panacea. FC-graphs require far more computational costs and may be prohibited in the case of tremendous entities. 

\subsection{Relations to Other GNN Bottlenecks}
Multiple papers uncover that GNNs may perform poorly on tasks that require long-range dependencies.  \textbf{Under-reaching}~\cite{barcelo2020logical} states the inability of a node to be aware of nodes that are farther away than the number of layers. This can be naively avoided by deepening GNNs, but the increase in layers can lead to a severe decline in prediction capability. \textbf{Over-smoothing}~\cite{li2018deeper,chen2020measuring} and \textbf{over-squashing}~\cite{alon2020bottleneck,topping2021understanding} are two mainstream accepted explanations for this decline. The former owes the failure to indistinguishable node representations when stacking too many layers, while the latter believes that the compression of information from the exponentially receptive field is the core reason for degraded performance in long-range problems. 

Our work discusses the representation bottleneck of GNNs, which highly resonates with but is essentially different from over-squashing. Over-squashing, the prime motivator for graph rewiring, is proposed by the information loss due to long-range dependencies that are not adequately captured by GNNs, which can also be regarded as middle- or high-order interactions that are not fully seized due to improper graph connectivity. 
In opposition, our representation bottleneck is built on the theory of multi-order interactions~\cite{grabisch1999axiomatic} while over-squashing replies on the message propagation of node features. This makes our representation bottleneck more general than {over-squashing}.

\section{Interaction Strength-based Graph Rewiring}
In order to approach ${J^{(m)}}^*$, recent work~\cite{deng2021discovering} imposes two losses to encourage or penalize learning interactions of specific complexities. Nevertheless, they require models to make accurate predictions on subgraphs. However, variable removal brings about the out-of-distribution (OOD) problem~\cite{wang2022deconfounding}, which can arbitrarily manipulate the outcome of GNNs and produce erroneous predictions~\cite{zugner2018adversarial}. More importantly, these losses are based on the assumption that the image class remains regardless of pixel removal. However, it is not rational to assume the stability of graph properties if we alter their components. 
In this work, instead of intervening in the loss, we rely on the modification of the GNNs' inductive bias to capture the most informative order $m^*$ of interactions and therefore reach ${J^{(m)}}^*$. 

Unfortunately, $m^*$ can never be known unless sufficient domain knowledge is supplied. As shown in Fig.~\ref{fig:str_order_cnn_mlp}, $J^{(m)}$ for CNNs does not directly dive into low-order of interactions in the initial training epochs (\emph{e.g.}, 10 or 50 epochs). Instead, CNNs have the inclination to learn a more informative order of interactions (\emph{e.g.}, middle-order) regardless of the inductive bias, which has also been observed for GNNs. So, motivated by this subtle tendency, we resort to the order of interactions that increase the most in $J^{(m)}$ during training as the guidance to reconstruct graphs and estimate ${J^{(m)}}^*$. 

To this end, we dynamically adjust the reception fields of each node within graphs by establishing or destroying edges based on the interaction strength $J^{(m)}$. To begin with, we randomly sample a mini-batch and calculate its corresponding interaction strengths. If the maximum increase of some order, \emph{i.e.}, $\max_m\left(\Delta J^{(m)}_t - J^{(m)}_{t-1}\right)$, exceeds the pre-defined threshold $\Bar{J}$, then we modify the number of nearest neighbors $K$ to be closer to the order $m^*$ whose interaction strength rises the most as $K = {(m^* + K)}/{2}$. After that, we reconstruct the node connectivity for $\{\mathcal{G}_i\}_{i=1}^N$ according to the new $K$. This process is iteratively implemented every $\Delta e$ epoch for an efficient training speed. 

Such a method is often generically referred to as \emph{graph rewiring}~\cite{topping2021understanding}, so it is dubbed Interaction Strength-based Graph Rewiring, dubbed \textbf{ISGR}, as described in Alg.~\ref{alg: alg}. By adjusting the graph topology that arouses the inductive bias, GNNs are enabled to break the representation bottleneck to some extent, and $J^{(m)}$ is able to gradually approximate ${J^{(m)}}^*$. 

\begin{algorithm}[ht]
    \caption{The workflow of the ISGR Algorithm.}
    \begin{algorithmic}
    \STATE {\bfseries Input:}  threshold $\Bar{J}$, total epochs $E$, interval $\Delta e$
    \FOR{$t = 1, 2, ..., {E}/{\Delta e}$}
        \STATE $J^{(m)}_t\leftarrow$ calculate Equ.~\ref{equ: multi_order_interaction} on a random batch
        \IF{$\max_m \left(J^{(m)}_t - J^{(m)}_{t-1} \right) > \Bar{J}$} 
            \STATE $m^* \leftarrow \mathrm{argmax}_m \left(J^{(m)}_t - J^{(m)}_{t-1} \right)$ 
            \STATE $K\leftarrow \frac{m^* + K}{2}$ $\qquad\triangleright$ make $K$ closer to $m^*$
            \STATE $\{\mathcal{G}_i\}_{i=1}^N \leftarrow K$  $\:\;\quad\triangleright$ reset $K$ for all KNN-graphs 
        \ENDIF
    \ENDFOR
    \end{algorithmic}
    \label{alg: alg}
\end{algorithm}

\section{Experiments: Effects of ISGR Algorithm}
\label{sec:exp}
\subsection{Tasks and Datasets}
We present four tasks to explore the representation patterns of GNNs for scientific research. Among them, molecular property prediction and Hamiltonian dynamics are graph-level prediction tasks, while Newtonian dynamics and molecular dynamics simulations are node-level ones. 

\paragraph{Molecular property prediction} Forecasting a broad range of molecular properties is a fundamental task in drug discovery. The acceleration of finding better drug candidates is compelling since the average cost for a new drug is at a sky-high price, where DL methods, especially GNNs, play an irreplaceable role. 
The properties in current molecular collections can be mainly divided into four categories: quantum mechanics, physical chemistry, biophysics, and physiology, ranging from molecular-level properties to macroscopic influences on the human body~\cite{wu2018moleculenet}. We use two benchmark datasets. QM7~\cite{blum2009970} is a subset of GDB-13 and is composed of 7K molecules. QM8~\cite{ramakrishnan2015electronic} is a subset of GDB-17 with 22K molecules. Note that QM7 and QM8 provide one and twelve properties, respectively, and we merely use the \emph{E1-CC2} property in QM8 for simplicity. 

\paragraph{Newtonian dynamics} Newtonian dynamics describes the dynamics of particles according to Newton's law of motion: the motion of each particle is modeled using incident forces from nearby particles, which change its position, velocity, and acceleration. Several important forces in physics, such as the gravitational force, are defined on pairs of particles, analogous to the message function of GNNs. We adopt the N-body particle simulation dataset~\cite{cranmer2020discovering}. It consists of N-body particles under six different interaction laws.
The following six forces are utilized in this dataset of Newtonian dynamics: (1) $1 / r$ orbital force: $-m_{1} m_{2} \hat{r} / r$; (2) $1 / r^{2}$ orbital force $-m_{1} m_{2} \hat{r} / r^{2}$; (3) charged particles force $q_{1} q_{2} \hat{r} / r^{2}$; (4) damped springs with $|r-1|^{2}$ potential and damping proportional and opposite to speed; (5) discontinuous forces, $-\left\{0, r^{2}\right\} \hat{r}$, switching to 0 force for $r<2$; and (6) springs between all particles, a $(r-1)^{2}$ potential. There, we only use the spring force for our experiments.

\paragraph{Hamiltonian dynamics} Hamiltonian dynamics~\cite{greydanus2019hamiltonian} describes a system's total energy $\mathcal{H}(\mathbf{q},\mathbf{p})$ as a function of its canonical coordinates $\mathbf{q}$ and momenta $\mathbf{p}$, \emph{e.g.}, each particles' position and momentum. The dynamics of the system change perpendicularly to the gradient of $\mathcal{H}$: $\frac{\mathrm{d} \mathbf{q}}{\mathrm{d} t}=\frac{\partial \mathcal{H}}{\partial \mathbf{p}}, \frac{\mathrm{d} \mathbf{p}}{\mathrm{d} t}=-\frac{\mathrm{d} \mathcal{H}}{\mathrm{d} \mathbf{q}}$. There, we take advantage of the same datasets from the Newtonian dynamics case study and attempt to learn the scalar total energy $\mathcal{H}$ of the system.

\paragraph{Molecular dynamics simulations} MD has long been the \emph{de facto} choice for modeling complex atomistic systems from first principles. 
There, MD simulations are carried out using the standard quantum chemistry computational method, density functional theory (DFT), which is different from the classic force field in Newtonian dynamics. 
We adopt ISO17~\cite{schutt2018schnet}, which has 129 molecules and is generated from MD simulations using the Fritz-Haber Institute \emph{ab initio} simulation package~\cite{blum2009ab}. Each molecule contains 5K conformational geometries and total energies with a resolution of 1 femtosecond in the trajectories. We predict the atomic forces in molecules at different timeframes. The molecules are randomly drawn from the largest set of isomers in QM9~\cite{ramakrishnan2014quantum} with a fixed composition of atoms ($C_7O_2H_{10}$) arranged in different chemically valid structures.  

\subsection{Experimental Settings}
\label{sec: exp_detail}
Two state-of-the-art geometric GNNs are selected to perform on these two types of graphs. We pick up Equivariant GNN (EGNN)~\cite{satorras2021n} for KNN-graphs, and {Molformer}~\cite{wu20213d} with no motifs for FC-graphs. EGNN is a roto-translation and reflection equivariant without the spherical harmonics. Molformer is a variant of Transformer~\cite{vaswani2017attention}, designed for molecular graph learning.

\paragraph{Baselines}
We compare ISGR to various graph rewiring methods. \textbf{+FA}~\cite{alon2020bottleneck} modifies the last GNN layer to be fully connected. \textbf{DIGL}~\cite{klicpera2019diffusion} leverages generalized graph diffusion to smooth out the graph adjacency and promote connections among nodes at short diffusion distances. \textbf{SDRF}~\cite{topping2021understanding} is the state-of-the-art rewiring technique and alleviates a graph's strongly negatively curved edges.

\paragraph{Training Details}
All experiments are implemented by Pytorch on an A100 GPU. An Adam~\cite{kingma2014adam} optimizer is used without weight decay, and a ReduceLROnPlateau scheduler is enforced to adjust it with a factor of 0.6 and patience of 10. The initial learning rate is 1e-4, and the minimum learning rate is 5e-6. The batch size is 512 for the sake of a fast training speed. Each model is trained for 1200 epochs, and early stopping is used if the validation error fails to decrease for 30 successive epochs. We randomly split each dataset into training, validation, and test sets with a ratio of 80/10/10. 

For both EGNN and Molformer, the number of layers (\emph{i.e.}, depths) is 3, and the dimensions of the input feature are 32. Besides, Molformer has 4 attention heads and a dropout rate of 0.1. The dimension of the feed-forward network is 128. It is worth noting that we employ multi-scale self-attention with a distance bar of $[0.8, 1.6, 3]$ to achieve better performance. This multi-scale mechanism helps Molformer to concentrate more on local contexts. However, it does not harm FC-graphs, and the connections between all pairs of entities remain. We also discover that the multi-scale mechanism has little impact on the distribution of $J^{(m)}$ and $J^{(m)}_D$. Regarding the setup of the ISGR algorithm, the threshold $\Bar{J}$ to adjust the number of neighbors is tuned via a grid search. The interval of epochs $\Delta e$ is 10, and the initial $k_0=8$. Concerning baselines, we follow~\cite{klicpera2019diffusion} and~\cite{topping2021understanding} and optimize hyperparameters by random search. Table~\ref{tab: hyper_digl} documents $\alpha$, $k$, and $\epsilon$ for DIGL, whose descriptions can be found in~\cite{klicpera2019diffusion}. Table~\ref{tab: hyper_sdrf} reports the maximum iterations, $\tau$ and $C^+$ for SDRF, whose descriptions is available in~\cite{topping2021understanding}.

\begin{table*}[ht] 
\caption{Hyperparameters for DIGL as the baseline algorithm.}
\label{tab: hyper_digl} 
\centering
\resizebox{1.8\columnwidth}{!}{%
\begin{tabular}{@{}c| cc | cc | cc | cc | cc } \toprule
    Task & \multicolumn{2}{c|}{{Newtonian Dynamics}} & \multicolumn{2}{c|}{Molecular Dynamics} &\multicolumn{2}{c|}{{Hamiltonian Dynamics}} & \multicolumn{2}{c|}{QM7} & \multicolumn{2}{c}{QM8} \\  
    Model & EGNN & Molformer & EGNN & Molformer & EGNN & Molformer  & EGNN & Molformer & EGNN & Molformer \\ \midrule
    $\alpha$ & 0.0259 & 0.1284 & 0.0732 & 0.1041 &  0.1561 & 0.3712 & 0.0655 & 0.2181 & 0.1033 & 0.1892 \\
    $k$ & 32 & 32 & 32 & 32 & 64 & 64 & - & - & - & - \\
    $\epsilon$ & - & - & - & - & 0.0001 & - & - & - & - & 0.0002\\ \bottomrule
\end{tabular}}
\end{table*}

\begin{table*}[ht]
\caption{Hyperparameters for SDRF as the baseline algorithm.}
\label{tab: hyper_sdrf} 
\centering
\resizebox{1.8\columnwidth}{!}{%
\begin{tabular}{@{}c| cc | cc | cc | cc | cc } \toprule
    Task & \multicolumn{2}{c|}{{Newtonian Dynamics}} & \multicolumn{2}{c|}{Molecular Dynamics} &\multicolumn{2}{c|}{{Hamiltonian Dynamics}} & \multicolumn{2}{c|}{QM7} & \multicolumn{2}{c}{QM8} \\ 
    Model & EGNN & Molformer & EGNN & Molformer & EGNN & Molformer  & EGNN & Molformer & EGNN & Molformer \\ \midrule
    Max Iter. & 15 & 11 & 39 & 34 & 16 & 13 & 22 & 17 &  25 & 12 \\
    $tau$ & 120 & 163 & 54 & 72 & 114 & 186 & 33 & 35 & 48 & 60  \\
    $C^+$ & 0.73 & 1.28 & 1.44 & 1.06 & 0.96 & 0.88 & 0.53 & 0.70 & 0.69 & 0.97\\ \bottomrule
\end{tabular}}
\end{table*}

We create a simulated system with 10 identical particles with a unit weight for Hamiltonian and Newtonian cases. For QM7, QM8, and ISO17 datasets, we sample 10 molecules that have the lowest MAE. For Hamiltonian and Newtonian datasets, we sample 100 timeframes with the lowest prediction errors. Then, for each molecule or dynamic system, we compute all pairs of entities $i,j\in [N]$ without any sampling strategy. Moreover, we limit the number of atoms between 10 and 18 to compute the interaction strengths for QM7 and MQ8.

\paragraph{Examination of the Normal Distribution Hypothesis}
We use \emph{scipy.stats.normaltest} in the Scipy package~\cite{virtanen2020scipy} to test the null hypothesis that $\frac{\partial \Delta f(i, j, S)}{\partial W}$ comes from a normal distribution, i.e, $\frac{\partial \Delta f(i, j, S)}{\partial W} \sim \mathcal{N}\left(0, \sigma^{2}\right)$. This test is based on D'Agostino and Pearson's examination and combines skew and kurtosis to produce an omnibus test of normality.  The $p$-values of well-trained EGNN and Molformer on the Hamiltonian dynamics dataset are 1.97147e-11 and 2.38755e-10, respectively. The $p$-values of randomly initialized EGNN and Molformer on the Hamiltonian dynamics dataset are 2.41749e-12 and 9.78953e-07, respectively. Therefore, we are highly confident in rejecting the null hypothesis (\emph{e.g.}, $\alpha=0.01$) and insist that $\frac{\partial \Delta f(i, j, S)}{\partial W}$ depends on the data distributions of downstream tasks and the backbone model architectures.

\paragraph{Node Features for $\mathcal{G} \backslash \mathcal{G}_S$}
Meanwhile, it also needs to be clarified how to compute the node features for $\mathcal{G} \backslash \mathcal{G}_S$. Noteworthily, the widely-used setting proposed by~\cite{ancona2019explaining} for sequences or pixels is inapplicable to GNNs,
In real-world scenarios, the most crucial feature of nodes (atoms or particles) is their classes (\emph{e.g.}, one-hot embeddings).
Because an average over graph instances can lead to an ambiguous node type. Alternatively, we consider dropping nodes and corresponding edges in $\mathcal{G} \backslash \mathcal{G}_S$ instead of replacing them with a mean value. 

\begin{table*}[t] 
\caption{Comparison of graph rewiring methods for graph-level and node-level prediction tasks. The best values are in bold, and the second-best values are underlined (this formatting applies to all subsequent tables).}
\label{tab: improve}
    \setlength{\tabcolsep}{0.3mm}  
    \centering
\resizebox{1.0\linewidth}{!}{%
\begin{tabular}{@{}l| cc | cc | cc | cc | cc} \toprule
    Task & \multicolumn{2}{c|}{{Hamiltonian Dynamics}} & \multicolumn{2}{c|}{QM7} & \multicolumn{2}{c|}{QM8} & \multicolumn{2}{c|}{{Newtonian Dynamics}} & \multicolumn{2}{c}{Molecular Dynamics} \\ 
    Model & EGNN & Molformer & EGNN & Molformer & EGNN & Molformer & EGNN & Molformer & EGNN & Molformer \\ \midrule
    None & 1.392$\, \pm \,$0.042 & 1.545$\, \pm \,$0.036 & 68.182$\, \pm \,$3.581 &  51.119$\, \pm \,$2.193  & 0.012$\, \pm \,$0.001 & 0.012$\, \pm \,$0.001 & 6.951 $\, \pm \,$ 0.098 &  1.929 $\, \pm \,$ 0.051 &  1.409 $\, \pm \,$ 0.082  &  0.848 $\, \pm \,$ 0.053 \\
    +FA & {1.168$\, \pm \,$0.043} & -- & \underline{55.288$\, \pm \,$3.074} &  --  & 0.012$\, \pm \,$0.001 & -- & \underline{5.348 $\, \pm \,$ 0.183} &  -- &  \underline{0.826 $\, \pm \,$ 0.105}  &  --  \\
    +DIGL & 1.151$\, \pm \,$0.044 & 1.337$\, \pm \,$0.072 & 61.028$\, \pm \,$3.804 &  41.188$\, \pm \,$5.329 & 0.012$\, \pm \,$0.001 & 0.011$\, \pm \,$0.001 & 5.637 $\, \pm \,$ 0.147 &  1.902 $\, \pm \,$ 0.081 &  1.108 $\, \pm \,$ 0.131  &  0.790 $\, \pm \,$ 0.078\\
    +SDRF & \underline{1.033$\, \pm \,$0.790} & \underline{1.265$\, \pm \,$0.039} & 59.921$\, \pm \,$3.765 & \underline{35.792$\, \pm \,$4.565} & \underline{0.011$\, \pm \,$0.001} & \underline{0.011$\, \pm \,$0.001} & 5.460 $\, \pm \,$ 0.133  &  \underline{1.885 $\, \pm \,$ 0.068} &  0.942 $\, \pm \,$ 0.152  &  \underline{0.751 $\, \pm \,$ 0.046} \\ \midrule
    +ISGR & \textbf{0.892$\, \pm \,$0.051} &  \textbf{1.250$\, \pm \,$0.029} &  \textbf{53.134$\, \pm \,$2.711}  &  \textbf{34.439$\, \pm \,$4.017}  &  \textbf{0.011$\, \pm \,$0.000}  &  \textbf{0.010$\, \pm \,$0.001} & \textbf{4.734 $\, \pm \,$ 0.103} &  \textbf{1.879 $\, \pm \,$ 0.066} &  \textbf{0.713 $\, \pm \,$ 0.097}  &  \textbf{0.736 $\, \pm \,$ 0.048} \\ \bottomrule
\end{tabular}}
\end{table*}

\begin{figure*}[t]
\centering
    \subfigbottomskip=-2pt
    \subfigcapskip =-2pt
    \subfigure[]{\includegraphics[width=0.17\linewidth]{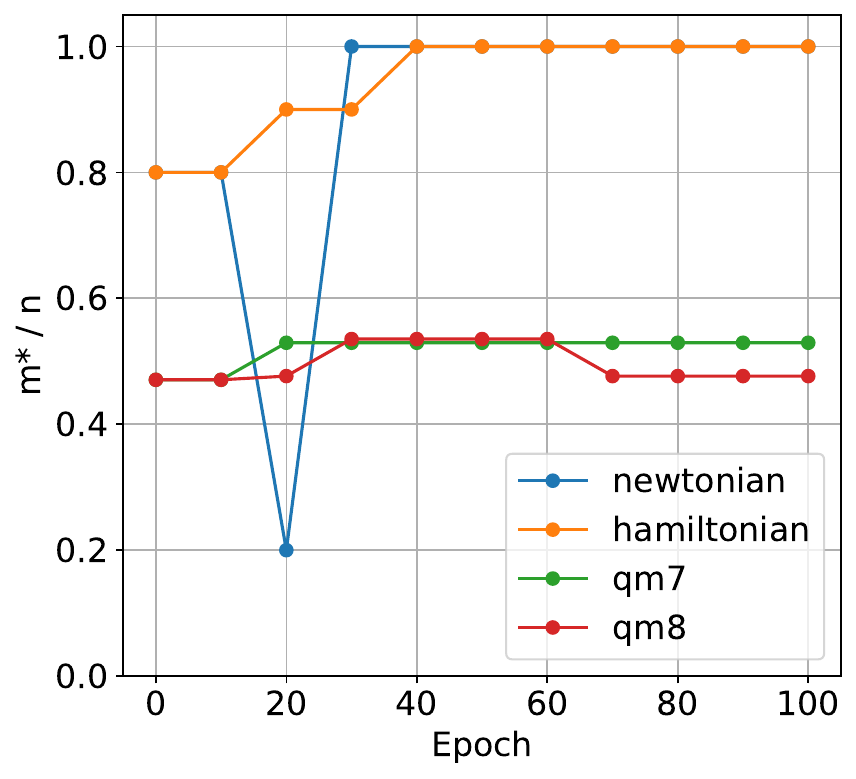} }
    \subfigbottomskip=-2pt
    \subfigcapskip =-4pt
    \subfigure[]{\includegraphics[width=0.80\linewidth]{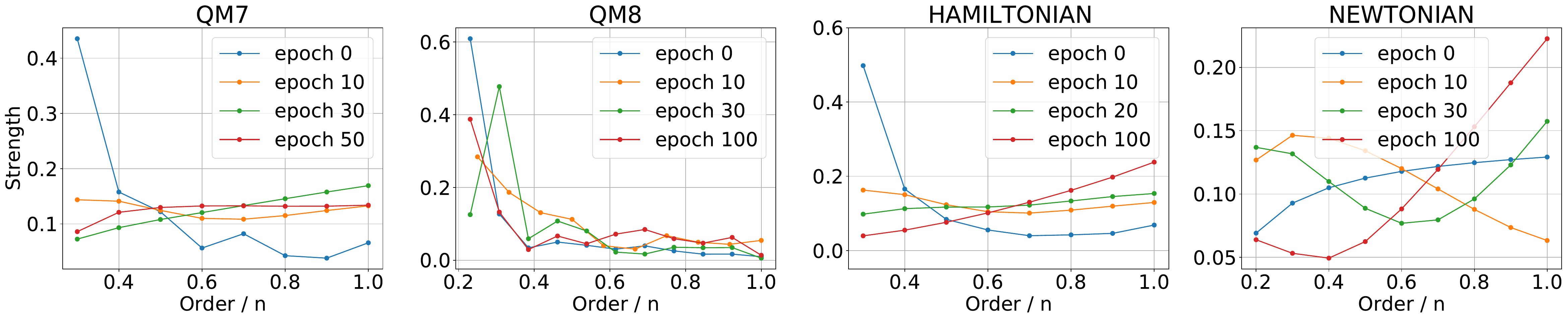} }
\caption{(a) The change of $m^*$ over epochs for EGNN on QM7, QM8, Hamiltonian, and Newtonian datasets. (b) Change in interaction strengths over epochs for EGNN on these four datasets.}
\label{fig: k_j_training}
\end{figure*}

\subsection{Comparison Results}
Extensive experiments are conducted to examine the efficiency of our ISGR method. Tab.~\ref{tab: improve} documents the outcome with the mean and standard deviation of three repetitions, where the top two are in bold and underlined, respectively. Our ISGR algorithm significantly improves the performance of EGNN and Molformer upon all baselines on both graph-level and node-level tasks. Particularly, the promotion of ISGR for EGNN is much higher, which confirms our assertion that GNNs based on KNN-graphs are more likely to suffer from bad inductive bias. On the flip side, the improvement for Molformer in QM7 is more considerable than in QM8. This proves that GNNs based on FC-graphs are more easily affected by inappropriate inductive bias (\emph{i.e.}, full connection) when the data is insufficient, since the size of QM7 (7K) is far smaller than QM8 (21K). We also see that +FA outweighs DIGL and SDRF, the rewiring algorithms by edge sampling, when the graph connectivity is built on KNN. However, when encountering FC-graphs, +FA loses efficacy, and SDRF achieves a larger improvement than DIGL.

\paragraph{Change of $m^*$}
Fig.~\ref{fig: k_j_training} plots the variation tendency of $m^*$ over epochs, showing that different tasks enjoy various optimal $K$ (denoted as $K^*$). Explicitly, Hamiltonian dynamics and Newtonian dynamics benefit from the full connection (${K^*}/{n} = 1$), while the molecular property prediction benefits more from middle-order interactions (${K^*}/{n} \approx 0.5$). This phenomenon perfectly fits the physical laws because the system in the Newtonian and Hamiltonian datasets is extremely compact with close pairwise distances. Those particles are more likely to be influenced by all the other nodes.

\paragraph{The change of interaction strengths during training}
Fig.~\ref{fig: k_j_training} also depicts how $J^{(m)}$ changes when the training proceeds with our ISGR algorithm. Although for molecular property prediction and Hamiltonian dynamics, $J^{(m)}_D$ mostly concentrate on low-order interactions ($m/n \leq 0.4$), $J^{(m)}$ progressively adjust to middle- and high-order ($m/n\geq 0.4$). Regarding Newtonian dynamics, $J^{(m)}_D$ is very smooth, but $J^{(m)}$ at initial epochs (\emph{i.e.}, $10$ and $20$ epochs) oddly focus on low-order interactions ($m/n\leq 0.4$). Nevertheless, our ISGR method timely corrects the wrong tendency, and eventually, $J^{(m)}$ becomes more intensive in segments of middle- and high-order ($m/n\geq 0.6$).

\subsection{More Experiments on Other Domains}
In addition to the promise shown in solving scientific problems, we include several commonly used graph datasets in other domains to demonstrate the broad applications of ISGR. Specifically, we use Cora~\cite{sen:aim08}, Citeseer~\cite{rossi:aaai15}, Pubmed~\cite{namata2012query}, Chameleon~\cite{pei2020geom}, and ogbn-papers100M~\cite{hu2020open}, and select GCN, GAT, GraphSAGE, and GIN as backbones. Cora, Citeseer, and PubMed are citation networks in which each publication is defined by a 0/1-valued word vector indicating the presence or absence of the corresponding dictionary word. Chameleon is a network based on English Wikipedia, representing pages dedicated to the topic of chameleon. Articles are represented by nodes, and mutual links between them are represented by edges. ogbn-papers100M is a paper citation network extracted from the Microsoft Academic Graph. The former contains 111 million papers while the latter only includes the Computer Science arXiv papers. A 128-dimensional feature vector is generated for each paper (node) in the dataset by averaging the embeddings of words in the title and abstract. Table~\ref{tab:graph} reports the mean and the standard deviation of accuracy on those five classification tasks. Our ISGR gains a significant improvement over baseline mechanisms. 
\begin{table*}[t] 
\caption{Extensive performance comparison of different graph rewiring methods for diverse classification datasets in other domains, where OOM represents the out-of-memory training. The number in the bracket is the P-value of the paired t-test. }
\label{tab:graph}
\centering
\resizebox{0.65\linewidth}{!}{%
\begin{tabular}{@{}l| cc cc c} \toprule
    Task & Cora & Citeseer &  Pubmed & Chameleon & ogbn-papers \\  \midrule
    \textbf{GCN} & 86.4$\, \pm \,$1.6 &  75.6$\, \pm \,$1.0 &  86.4$\, \pm \,$0.5  & 43.1$\, \pm \,$2.7  &   67.1$\, \pm \,$0.2   \\ 
    +FA & 87.0$\, \pm \,$1.2  &  76.2$\, \pm \,$0.7 &  86.8$\, \pm \,$0.4 & 43.6$\, \pm \,$2.1   & OOM   \\  
    +DIGL & 86.8$\, \pm \,$1.4 &  75.6$\, \pm \,$0.9 &  86.5$\, \pm \,$0.5 &  43.0$\, \pm \,$3.0   & 67.2$\, \pm \,$0.1   \\ 
    +SDRF & 88.2$\, \pm \,$1.4   &  76.8$\, \pm \,$1.0 & 87.1$\, \pm \,$0.5   & 50.3$\, \pm \,$3.6   & 67.5$\, \pm \,$0.1   \\ \midrule
    \multirow{2}{*}{+ISGR} & \textbf{89.4$\, \pm \,$1.5} &  \textbf{77.5$\, \pm \,$1.0} & \textbf{87.7$\, \pm \,$0.5} & \textbf{51.4$\, \pm \,$1.9} & \textbf{67.6$\, \pm \,$0.2}   \\
     &  (0.001)  & (0.002) & (0.024) & (0.007) & (0.181) \\ \midrule
    \textbf{GAT} & 86.1$\, \pm \,$1.5   &  75.5$\, \pm \,$1.0   & 86.2$\, \pm \,$0.4 &  45.9$\, \pm \,$2.3 &  68.4$\, \pm \,$0.1   \\ 
    +FA & 86.8$\, \pm \,$1.3   &  76.4$\, \pm \,$0.9 & 86.9$\, \pm \,$0.5  & 46.0$\, \pm \,$2.0   & OOM   \\  
    +DIGL & 86.4$\, \pm \,$1.5 &  75.6$\, \pm \,$1.2 &  87.4$\, \pm \,$0.7 &  47.7$\, \pm \,$2.4   & 68.4$\, \pm \,$0.2   \\ 
    +SDRF & 88.2$\, \pm \,$1.3   &  76.5$\, \pm \,$1.0   & 87.5$\, \pm \,$0.6   & 50.1$\, \pm \,$4.0   & 68.6$\, \pm \,$0.1   \\ \midrule
    \multirow{2}{*}{+ISGR} & \textbf{89.0$\, \pm \,$1.1} &  \textbf{77.1$\, \pm \,$0.9} & \textbf{88.0$\, \pm \,$0.6} & \textbf{51.7$\, \pm \,$1.4} & \textbf{69.0$\, \pm \,$0.2}   \\ 
     &  (0.017)  & (0.028) & (0.019) & (0.007) & (0.025) \\ \midrule
    \textbf{GrapphSAGE} & 86.5$\, \pm \,$1.4   &  76.1$\, \pm \,$0.9   & 87.5$\, \pm \,$0.4   & 47.7$\, \pm \,$2.0   & 68.2$\, \pm \,$0.1   \\ 
    +FA & 86.6$\, \pm \,$1.5   &  76.4$\, \pm \,$1.1   & 86.3$\, \pm \,$0.7   & 46.0$\, \pm \,$2.1   & OOM   \\  
    +DIGL & 86.5$\, \pm \,$1.5   &  76.2$\, \pm \,$1.1   & 87.4$\, \pm \,$0.7 &  47.7$\, \pm \,$2.4  & 68.2$\, \pm \,$0.0   \\ 
    +SDRF & 88.1$\, \pm \,$1.2   &  77.0$\, \pm \,$0.9   & 87.5$\, \pm \,$0.4   & 50.3$\, \pm \,$3.8   & 68.4$\, \pm \,$0.1   \\  \midrule
   \multirow{2}{*}{+ISGR} & \textbf{88.8$\, \pm \,$1.3} &  \textbf{77.5$\, \pm \,$1.0} & \textbf{88.4$\, \pm \,$0.5} & \textbf{52.0$\, \pm \,$1.7} & \textbf{68.6$\, \pm \,$0.0}    \\ 
     &  (0.020)  & (0.022) & (0.016) & (0.005) & (0.047) \\ \midrule
    \textbf{GIN} & 87.4$\, \pm \,$1.5   &  76.3$\, \pm \,$1.0   & 88.0$\, \pm \,$0.5   & 46.1$\, \pm \,$1.7   & 69.8$\, \pm \,$0.1   \\ 
    +FA & 87.8$\, \pm \,$1.1   &  76.6$\, \pm \,$1.0 &  88.2$\, \pm \,$0.6 & 46.5$\, \pm \,$2.1 & OOM   \\  
    +DIGL & 87.5$\, \pm \,$1.6   &  76.4$\, \pm \,$1.2  & 77.9$\, \pm \,$0.6   & 46.1$\, \pm \,$1.8   & 69.7$\, \pm \,$0.1   \\ 
    +SDRF & 87.7$\, \pm \,$1.3  &  77.1$\, \pm \,$1.0   & 88.5$\, \pm \,$0.5   & 50.4$\, \pm \,$3.9   & 70.3$\, \pm \,$0.1   \\ \midrule
    \multirow{2}{*}{+ISGR} & \textbf{89.8$\, \pm \,$1.8} &  \textbf{77.6$\, \pm \,$1.1} & \textbf{89.4$\, \pm \,$0.6} & \textbf{51.6$\, \pm \,$1.9} & \textbf{70.6$\, \pm \,$0.1}   \\ 
     &  (0.005)  & (0.037) & (0.008) & (0.002) & (0.061) \\ \bottomrule
\end{tabular}}
\end{table*}
\begin{table}[ht] 
\caption{Node classification results on Cora, Citeseer, Pubmed, where we report the mean and standard deviation of 10 runs with different random seeds. }
\label{tab:gsl}
\centering
\resizebox{0.95\linewidth}{!}{%
\begin{tabular}{@{}l| cc cc c} \toprule
    Task & Cora & Citeseer &  Pubmed \\  \midrule
    GCN & \underline{81.95$\, \pm \,$0.62} &  71.34$\, \pm \,$0.48 &  \underline{78.98$\, \pm \,$0.35}    \\ 
    SLAPS [2] & 72.29$\, \pm \,$1.01  &  70.00$\, \pm \,$1.29 &  70.96$\, \pm \,$0.99  \\  
    GEN [3] & 81.66$\, \pm \,$0.91 &  \textbf{73.21$\, \pm \,$0.62} &  78.49$\, \pm \,$3.98   \\ 
    Nodeformer [4] & 78.81$\, \pm \,$1.21  &  70.39$\, \pm \,$2.04  &  78.38$\, \pm \,$1.94   \\  \midrule 
    ISGR & \textbf{84.08$\, \pm \,$0.47} &  \underline{73.16$\, \pm \,$1.59} & \textbf{79.63$\, \pm \,$0.42}   \\ \bottomrule
\end{tabular}}
\end{table}

\subsection{Comparison with Graph Structure Learning Methods}
Graph structure learning (GSL) is an emerging subfield in graph representation learning, which aims to optimize the graph structure and the corresponding GNN representations jointly. By refining the graph structure, GSL can potentially empower GNNs to learn better representations with improved performance. GSL has been successfully applied to disease analysis and protein structure prediction. In this subsection, we conduct additional experiments to verify the efficacy of these GSL algorithms based on OpenGSL~\cite{zhou2023opengsl} at~\url{https://github.com/OpenGSL/OpenGSL}. We select SLAPS~\cite{fatemi2021slaps}, GEN~\cite{wang2021graph}, and NodeFormer~\cite{wu2022nodeformer} as baselines and test them on three mainstream graph learning datasets. The results are put in Table~\ref{tab:gsl}, which shows that ISGR outperforms others in two out of three benchmark datasets. More importantly, ISGR brings all the positive effects on GCN. In contrast, SLAPS, GEN, and NodeFormer can be worse than the vanilla GCN in PubMed. This verifies the potential of ISGR in mitigating the representation bottleneck issue more effectively than existing adaptive GSL methods.


\begin{figure}[ht]
\centering
    \includegraphics[width=0.9\linewidth]{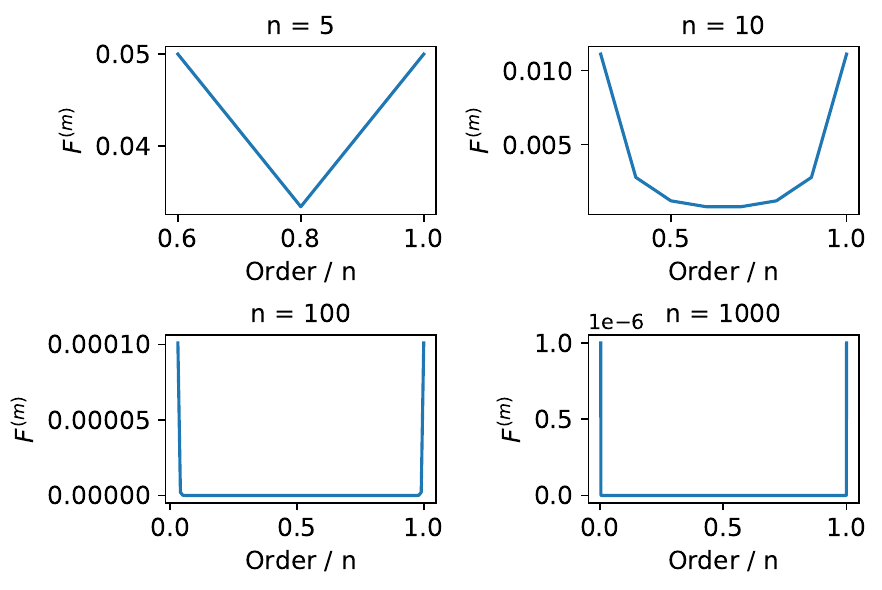}
    \vspace{-1.0em}
\caption{The theoretical distributions of $F^{(m)}$ under different $n$.}
    \label{fig:theoretical_order}
    \vspace{-1.0em}
\end{figure}

\section{Representation Bottlenecks of DNNs}
\label{sec: revisit}
\subsection{Revisiting Findings of Other DNNs}
Recently, the representation bottleneck of other DNNs (\emph{e.g.}, CNNs) has been increasingly investigated. 
Some~\cite{deng2021discovering} leverage $\Delta W^{(m)}(i, j)=R^{(m)} \frac{\partial I^{(m)}(i, j)}{\partial W}$ in Equ.~\ref{equ: delta_W} to represent the compositional component of $\Delta W$ w.r.t. ${\partial I^{(m)}(i, j)}/{\partial W}$. They claim that $\mid \Delta W^{(m)}(i, j)\mid$ is proportional to $F^{(m)}=\frac{n-m+1}{n(n-1)} / \sqrt{\binom{n-2}{m-2}}$ and experimentally show that $J^{(m)}$ of low-order (\emph{e.g.}, $m=0.05n$) is much higher than high-order (\emph{e.g.}, $m=0.95n$). 

Despite their fruitful progress, we argue that $F^{(m)}$ ought to be approximately the same when $m/n\rightarrow 0$ or $m/n\rightarrow 1$, as shown in Fig.~\ref{fig:theoretical_order}) if the empty set $\emptyset$ is excluded from the input for DNNs. In the Appendix, we demonstrate that even if $f(\emptyset)$ is taken into account and $n$ is large (\emph{e.g.}, $n\geq 100$), $J^{(m)}$ should be non-zero only when $m/n\rightarrow 0$. Therefore, if $\mid \Delta W^{(m)}(i, j)\mid$ is entirely dependent on $F^{(m)}$, DNNs should not capture middle- or high-order interactions. But DNNs have performed well in tasks that require high-order interactions such as protein-protein rigid-body docking and protein interface prediction~\cite{liu2020deep}. 

The inaccurate statement that $F^{(m)}$ determines $\mid \Delta W^{(m)}(i, j)\mid$ is due to the flawed hypothesis that the derivatives of $\Delta f(i, j, S)$ over model parameters, \emph{i.e.}, ${\partial \Delta f(i, j, S)}/{\partial W}$, conform to normal distributions (see the Appendix). 
${\partial \Delta f(i, j, S)}/{\partial W}$, undoubtedly, varies with contextual complexities (\emph{i.e.}, $|S|$), and replies on not only the data distribution of interaction strengths in particular datasets $J^{(m)}_D$ but the model architecture $f$. 

\begin{figure*}[t]
\centering
    \subfigbottomskip=-2pt
    \subfigcapskip =-6pt
    \subfigure[ResNet]{\includegraphics[width=0.47\linewidth]{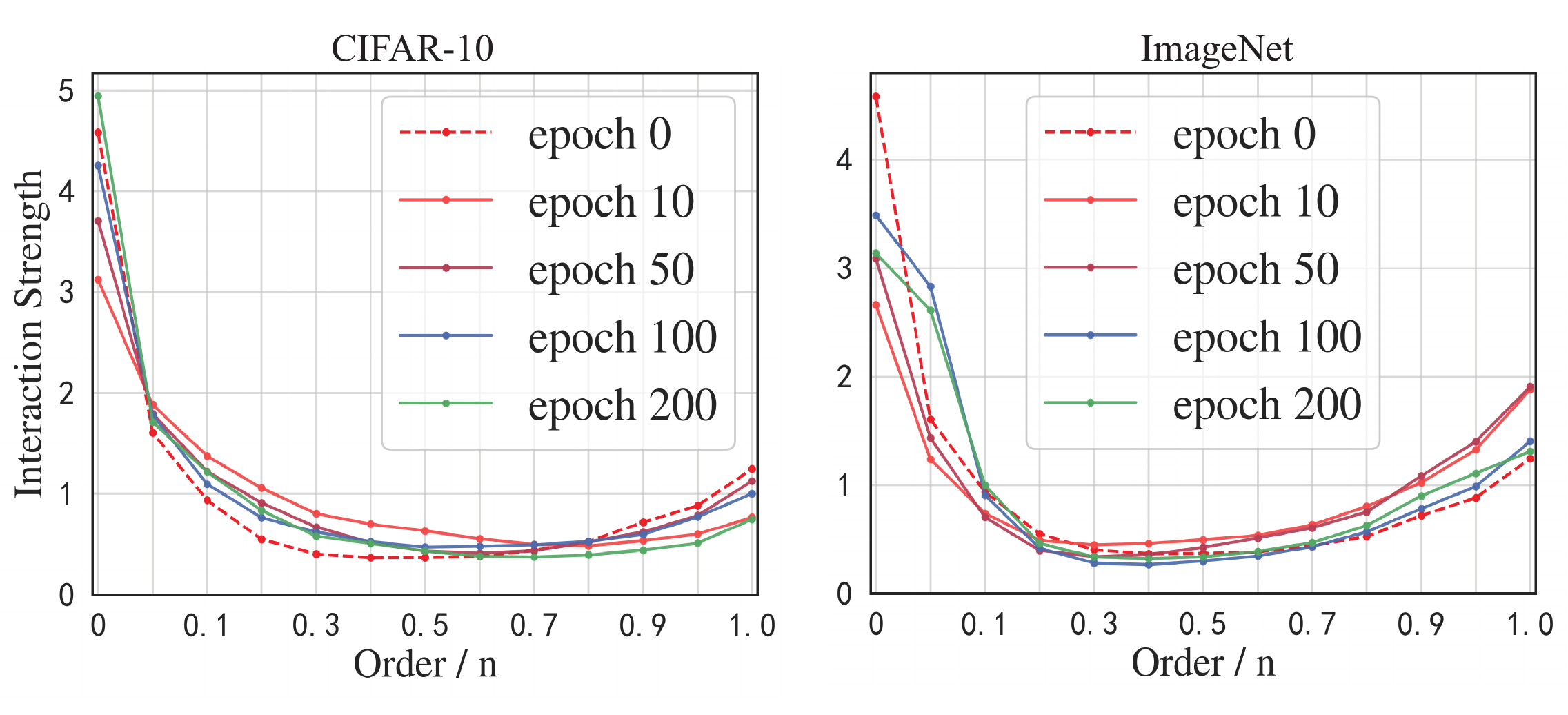}}
    \subfigure[MLP-Mixer]{\includegraphics[width=0.47\linewidth]{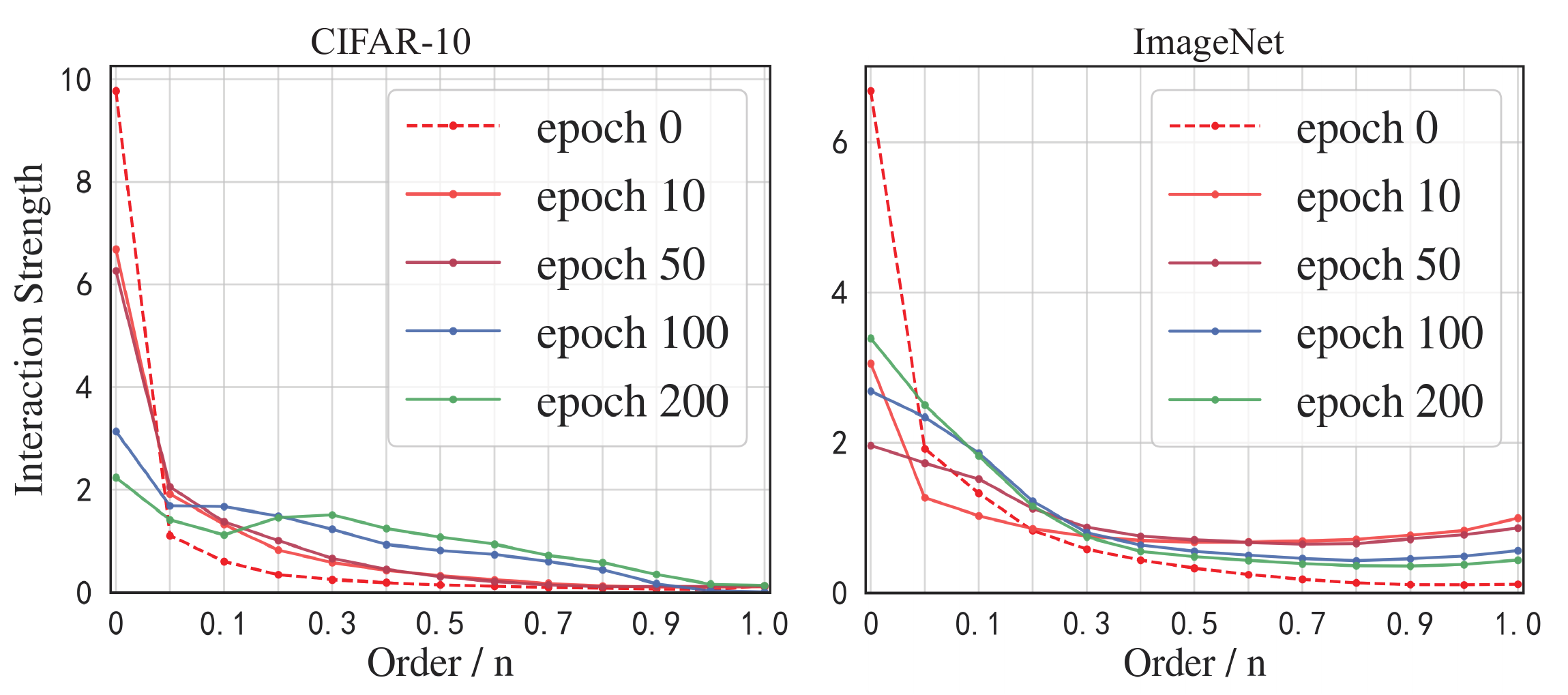}}
\caption{(a) The change of interaction strengths during training for ResNet on CIFAR-10 and ImageNet. (b) Change in interaction strengths during training for MLP-Mixer on CIFAR-10 and ImageNet.}
    \label{fig:str_order_cnn_mlp}
\end{figure*}

\subsection{Inductive Bias in Other DNNs}
To verify our dissent, we reproduce the interaction strengths of ResNet in CIFAR-10~\cite{krizhevsky2009learning} and ImageNet~\cite{russakovsky2015imagenet} in Fig.~\ref{fig:str_order_cnn_mlp}. The plot implies that $J^{(m)}_D$'s (referring to the epoch-0 curve) interactions of low and high orders are much stronger than middle orders. As previously analyzed, this phenomenon is due to a critical inductive bias of CNNs~\cite{he2016deep,han2021demystifying,icml2021deit,das2021classification}, \emph{i.e.}, locality. It assumes that entities are in spatially close proximity with one another and isolated from distant ones~\cite{battaglia2018relational}. Hence, CNNs with small kernel sizes are bound to low-order interactions. 

Several recent studies have shown that increasing the size of the kernel can alleviate the local inductive bias~\cite{ding2022scaling}. Based on their revelation, we examine the change of interaction strengths for MLP using MLP-Mixer~\cite{nips2021mlpmixer} in Fig.~\ref{fig:str_order_cnn_mlp}. Although MLP shares a similar $J^{(m)}_D$ with ResNet, its $J^{(m)}$ is much smoother. This is because MLP-Mixer assumes full connection of different patches with no constraint of locality. Therefore, it can learn a more adorable $J^{(m)}$. The implementation details on visual tasks are elaborated in the Appendix.


\subsection{Comparison between CNNs and GNNs}
Recall that in Fig.~\ref{fig: str_order}, $J^{(m)}$ learned by GNNs can deviate from and is nearly independent of the data distribution $J^{(m)}_D$. Precisely, $J^{(m)}_D$ for molecular property prediction in QM8 is more intensive on low orders. But after sufficient training, $J^{(m)}$ for EGNN mainly has high values for middle orders. $J^{(m)}$ for Molformer also increases the most in the middle-order segment. 
The trend of $J^{(m)}$ illustrates that subgraphs with a middle size are exceedingly informative substructures that reveal small molecules' biological or chemical properties. This finding verifies that motifs such as functional groups heavily determine molecular attributes~\cite{yu2020graph,wang2021towards,wu2022discovering}.
Similarly, although $J^{(m)}_D$ in QM7 concentrate on low-order, its $J^{(m)}$ are mainly allocated on middle-order. Especially for EGNN, its spike of $J^{(m)}$ is at $m=9$. Concerning Molformer, the segment of its $J^{(m)}$ that increases most is dispersive between $0.3\leq m/n \leq 0.5$.
While for Hamiltonian dynamic systems, $J^{(m)}_D$ is majorly intense for low and middle orders. In spite of that, $J^{(m)}$ of EGNN concentrates more on high orders but neglects low orders. Regarding node-level prediction tasks, the scenery is more straightforward. $J^{(m)}_D$ for EGNN and Molformer are in different shapes, but $J^{(m)}$ both move towards low orders for MD and high-order interactions for Newtonian dynamics. All those phenomenons bolster a considerable discrepancy between $J^{(m)}$ and $J^{(m)}_D$ for geometric GNNs. 

In opposition, as shown in Fig.~\ref{fig:str_order_cnn_mlp}, little difference exists between $J^{(m)}_D$ and $J^{(m)}$ (referring to non-zero epoch curves) at different epochs for both ResNet and MLP-Mixer. That is, $J^{(m)}$ learned by other sorts of DNNs, such as CNNs, are seldom divergent from $J^{(m)}_D$. This fact supports our previous claim in Sec.~\ref{sec: rep_gnns} that graph learning problems are more affected by inductive bias than visual tasks.

\section{Conclusions and Future Works}
\label{sec:conclusion}
\paragraph{Conclusions}
We discover and strictly analyze the representation bottleneck of GNNs from the complexity of interactions encoded in networks. Remarkably, inductive bias rather than the data distribution is more dominant in the expressions of GNNs to capture pairwise interactions, contradicting the behavior of other DNNs. 
Besides, empirical results demonstrate that inductive biases introduced by most graph construction mechanisms, such as KNN and full connection, can be sub-optimal and prevent GNNs from learning the most revelatory order of interactions. Inspired by this gap, we design a novel rewiring method based on the inclination of GNNs to encode more informative interactions. Extensive experiments on four synthetic and real-world tasks verify that our algorithm efficiently enables GNNs to approach the global minimum loss and break the representation bottleneck. 
We believe that this work offers a novel explanation for a well-known belief that subgraphs (e.g., motifs) contribute to recognizing graph properties.

\paragraph{Limitations and Future Directions}
Despite the reported success of our ISGR, GNNs still face limitations. In real-world applications like social networks~\cite{hamilton2017inductive,fan2019graph}, improper modeling of interactions may lead to the failure to classify fake users and posts. In addition, since GNNs have been extensively deployed in assisting biologists with the discovery of new drugs, inappropriate modeling can postpone the screening progress and impose negative consequences on the drug design. Another limitation of our work is that results are only demonstrated on small systems with few particles and two sorts of GNNs. The generalization of our claim remains to be investigated on larger datasets and more advanced architectures.

\bibliographystyle{IEEEtran}
\bibliography{cite}

\newpage
\begin{wrapfigure}{l}{0.3\columnwidth}
  \begin{center}
    \includegraphics[width=0.3\columnwidth]{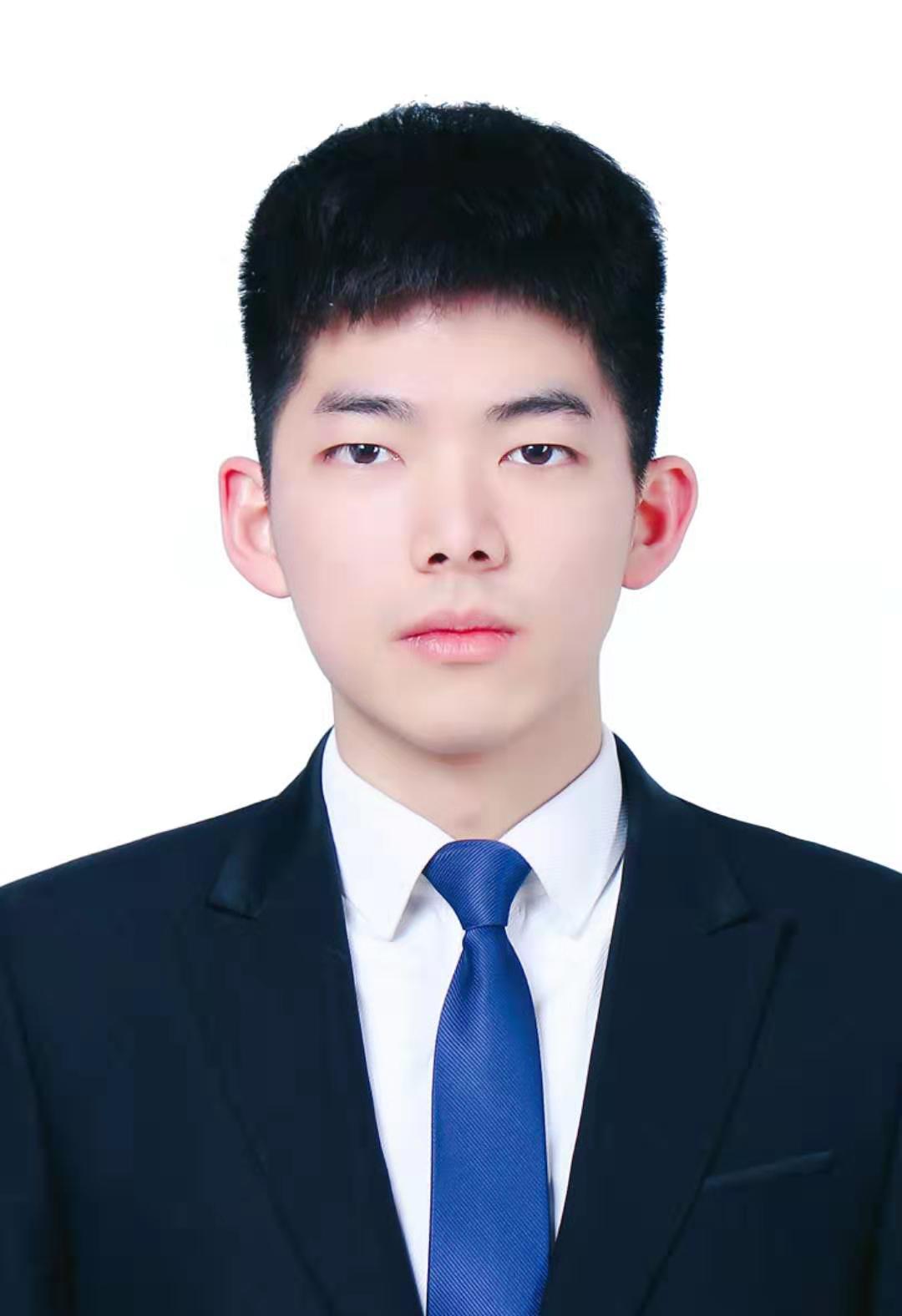}
  \end{center}
\end{wrapfigure}
\textbf{Fang Wu} is a CS Ph.D. student at Stanford University. He worked as a research assistant at Westlake University from 2021 to 2022, supervised by the chair Professor Stan Z. Li. His research interests include graph representation learning, geometric deep learning, and AI for science. He has published several first-author papers in Nature Communications, Advanced Science, Patterns, Communications Biology, ICML, NeurIPS, AAAI, IJCAI, etc. 
\newline\newline

\begin{wrapfigure}{l}{0.3\columnwidth}
  \begin{center}
    \includegraphics[width=0.3\columnwidth]{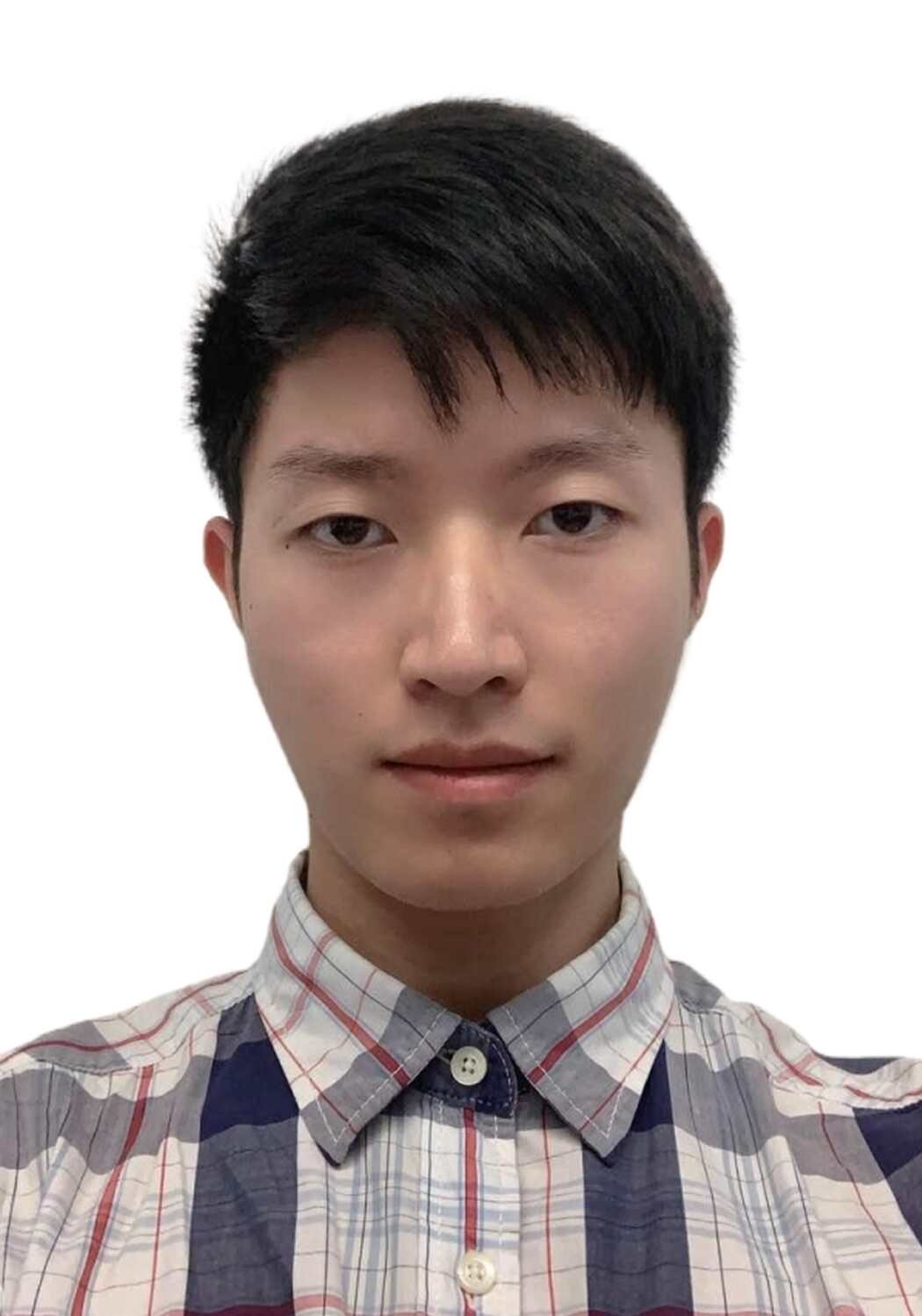}
  \end{center}
\end{wrapfigure}
\textbf{Siyuan Li} received the BEng. degree from Nanjing University. He is currently working toward a joint PhD degree at Zhejiang University and Westlake University, supervised by the chair Professor Stan Z. Li. His research interests include self-supervised learning, semi-supervised learning, manifold learning, and AI for science. He has published several papers in top-class AI conferences such as ICLR, ICML, NeurIPS, IJCAI, CVPR, ECCV, etc.
\newline\newline

\begin{wrapfigure}{l}{0.3\columnwidth}
  \begin{center}
    \includegraphics[width=0.3\columnwidth]{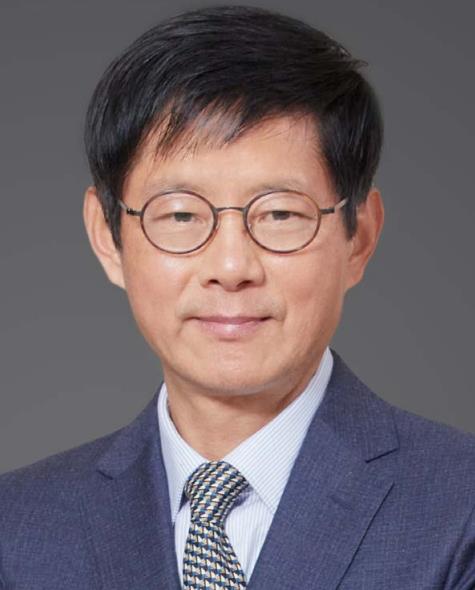}
  \end{center}
\end{wrapfigure} \textbf{Stan Z. Li} (Fellow, IEEE) received a BEng degree from Hunan University, an MEng degree from the National University of Defense Technology, and a PhD degree from Surrey University. He is currently a chair professor and the director of the Center for Artificial Intelligence Research and Innovation, School of Engineering, Westlake University. Before that, he is the director of the Center for Biometrics and Security Research, Institute of Automation, Chinese Academy of Sciences. He was a researcher lead at Microsoft Research Asia from 2000 to 2004. Before that, he was an associate professor at the Nanyang Technological University. His research interests include pattern recognition and machine learning, image and vision processing, face recognition, biometrics, and intelligent video surveillance. He has published more than 400 papers in international journals and conferences and authored and edited eight books. He was an associate editor of the IEEE Transactions on Pattern Analysis and Machine Intelligence and is acting as the editor-in-chief for the Encyclopedia of Biometrics. He served as a program co-chair for the International Conference on Biometrics in 2007, 2009, 2013, 2014, 2015, 2016, and 2018. He was elevated to IEEE fellow for his contributions to face recognition, pattern recognition, and computer vision, and is a member of the IEEE Computer Society.


\section*{Introduction and Theoretical Analysis of Multi-order Interactions}
\label{app: intro_multi_order}
\subsection{Introduction of Multi-order Interactions}
In this subsection, we give a more detailed introduction to the multi-order interaction, which is employed to analyze the representation ability of GNNs and CNNs in the main body, in case some audiences may find it hard to follow. This introduction uses previous studies~\cite{zhang2020interpreting,deng2021discovering} extensively as reference, and we strongly recommend that interested readers take a look at these articles. 

Here, we follow the same mathematical description of multi-order interactions as \cite{deng2021discovering} for better alignment and convenient comparison. $f$ is a pre-trained DNN (\emph{e.g.}, CNN, RNN, Transformer, GNN), and an input sample (\emph{e.g.}, an image or a text) has $n$ variables (\emph{e.g.}, pixels or words) denoted as $N=\{1,...,n\}$. Then, the original multi-order interaction between input variables is originally defined as follows:
\begin{equation}
    I^{(m)}(i, j)=\mathbb{E}_{S \subseteq N\backslash \{i,j\}, \mid S \mid=m}[\Delta f(i, j, S)], 3\leq m\leq n,
\end{equation}
where $\Delta f(i, j, S)=f(S\cup\{i, j\})-f(S \cup\{i\})-f(S \cup\{j\})+f(S)$ and $S\subset N$ represents the context with $m$ variables. $I^{(m)}(i, j)$ denotes the interaction between variables $i, j \in N$ of the $m$-th order, which measures the average interaction utility between $i, j$ under contexts of $m$ variables. 
There are five desirable properties that $I^{(m)}(i, j)$ satisfies:

\textbf{- Linear property.} If two independent games $f_1$ and $f_2$ are combined, obtaining $g(S)=f_1(S)+f_2(S)$, then the multi-order interaction of the combined game is equivalent to the sum of multi-order interactions derived from $f_1$ and $f_2$, \emph{i.e.},$I_{g}^{(m)}(i, j)=I_{f_1}^{(m)}(i, j)+I_{f_2}^{(m)}(i, j)$.

\textbf{- Nullity property.} If a dummy variable $i \in N$ satisfies $\forall S \subseteq N \backslash\{i\}, f(S \cup\{i\})=f(S)+f(\{i\})$, then variable $i$ does not interact with other variables, \emph{i.e.}, $\forall m, \forall j \in N \backslash\{i\}, I^{(m)}(i, j)=0$.

\textbf{- Commutativity property.} Intuitively, $\forall i, j \in N, I^{(m)}(i, j)=I^{(m)}(j, i)$.

\textbf{- Symmetry property.} Suppose two variables $i, j$ are equal in the sense that $i, j$ have same cooperations with other variables,  \emph{i.e.}, $\forall S \subseteq N \backslash\{i, j\}, f(S \cup\{i\})=f(S \cup\{j\})$, then we have $\forall k \in N, I^{(m)}(i, k)=I^{(m)}(j, k)$.

\textbf{- Efficiency property~\cite{deng2021discovering}.} The output of a DNN can be decomposed into the sum of interactions of different orders between different pairs of variables as:
\begin{equation}
    f(N)-f(\emptyset)=\sum_{i \in N} \mu_{i}+\sum_{i, j \in N, i \neq j} \sum_{m=0}^{n-2} w^{(m)} I^{(m)}(i, j),  
\end{equation}
where $\mu_{i}=f(\{i\})-f(\emptyset)$ represents the independent effect of variable $i$, and $w^{(m)}=\frac{n-1-m}{n(n-1)}$.

\paragraph{Connection with Shapley value and Shapley interaction index.}
Shapley value is introduced to measure the numerical importance of each player to the total reward in a cooperative game, which has been widely accepted to interpret the decision of DNNs in recent years~\cite{lundberg2017unified,ancona2019explaining}. For a given DNN and an input sample with a set of input variables $N=\{1, \ldots, n\}$, we use $2^{N}=\{S \mid S \subseteq$ $N\}$ to denote all possible subsets of variables of $N$. Then, DNN $f$ can be considered as $f: 2^{N} \rightarrow \mathbb{R}$ that calculates the output $f(S)$ of each specific subset $S \subseteq N$. Each input variable $i$ is regarded as a player, and the network output $f(N)$ of all input variables can be considered as the total reward of the game. The Shapley value aims to fairly distribute the network output to each variable as follows:
\begin{equation}
    \phi_{i}=\sum_{S \subseteq N \backslash\{i\}} \frac{\mid S \mid !(n-\mid S \mid-1) !}{n !}[f(S \cup\{i\})-f(S)],
\end{equation}
where $f(S)$ denotes the network output when we keep variables in $S$ unchanged while masking variables in $N \backslash S$ by following the setting in~\cite{ancona2019explaining}. It has been proven that the Shapely value is a unique method to fairly allocate overall reward to each player that satisfies \emph{linearity}, \emph{nullity}, \emph{symmetry}, and \emph{efficiency} properties.

\paragraph{Connections between the Shapley interaction index and the Shapley value.} Input variables of a DNN usually interact with each other rather than working individually. Based on the Shapley value,~\cite{grabisch1999axiomatic} further proposes the Shapley interaction index to measure the interaction utility between input variables. The Shapley interaction index is the only axiomatic extension of the Shapley value, which satisfies \emph{linearity}, \emph{nullity}, \emph{symmetry}, and \emph{recursive} properties. For two variables $i, j \in N$, the Shapley interaction index $I(i, j)$ can be considered as the change of the numerical importance of variable $i$ by the presence or absence of variable $j$.
\begin{equation}
    I(i, j)=\tilde{\phi}(i)_{j \text{ always present}}-\tilde{\phi}(i)_{j \text{ always absent}},
\end{equation}
where $\tilde{\phi}(i)_{j \text{always present}}$  denotes the Shapley value of the variable $i$ computed under the specific condition that variable $j$ is always present. $\tilde{\phi}(i)_{j \text { always absent}}$ is computed under the specific condition that $j$ is always absent.

\paragraph{Connections between the multi-order interaction and the Shapley interaction index.} Based on the Shapley interaction index,~\cite{zhang2020interpreting} further defines the order of interaction, which represents the contextual complexity of the interactions. It has been proven that the above Shapley interaction index $I(i, j)$ between variables $i, j$ can be decomposed into multi-order interactions as follows:
\begin{equation}
    I(i, j)=\frac{1}{n-1} \sum_{m=0}^{n-2} I^{(m)}(i, j).
\end{equation}

\subsection{Proof of Multi-order Interactions}
\label{app: I_m}
There we explain why $I^{(m)}(i, j)$ has no difference whether we include variables $\{i, j\}$ in $S$ or not. In the setting of~\cite{zhang2020interpreting,deng2021discovering}, $I^{(m)}(i, j)$ takes the following form:
\begin{equation}
    I^{(m)}(i, j)=\mathbb{E}_{S \subseteq N \backslash\{i, j\},\mid S \mid=m}[\Delta f(i, j, S)],
\end{equation}
where $\Delta f(i, j, S)=f(S \cup\{i, j\})-f(S \cup\{i\})-f(S \cup\{j\})+f(S)$ and $i, j\notin S$. While in our formulation, the order $m' = m + 2$ corresponds to the context $S'=S\cup \{i,j\}$. Now we denote our version of the multi-order interaction as $I'^{(m')}(i, j)$ with $\Delta' f(i, j, S)$ and aim to show that $I^{(m)}(i, j)=I'^{(m')}(i, j)$. 

It is trivial to obtain that $f(S \cup\{i, j\})-f(S \cup\{i\})-f(S \cup\{j\})+f(S) = f(S')-f(S' \backslash\{i\})-f(S' \backslash\{j\})+f(S'\backslash \{i, j\})$, which indicates that $\Delta f(i, j, S) = \Delta' f(i, j, S')$. Therefor, we can get $I^{(m \,+ \,2)}(i, j) =I'^{(m')}(i, j)$.  

\subsection{Theoretical Distributions of $F^{(m)}$}
\label{app: F(m)}
\begin{figure}[ht]
\centering
\includegraphics[scale=0.5]{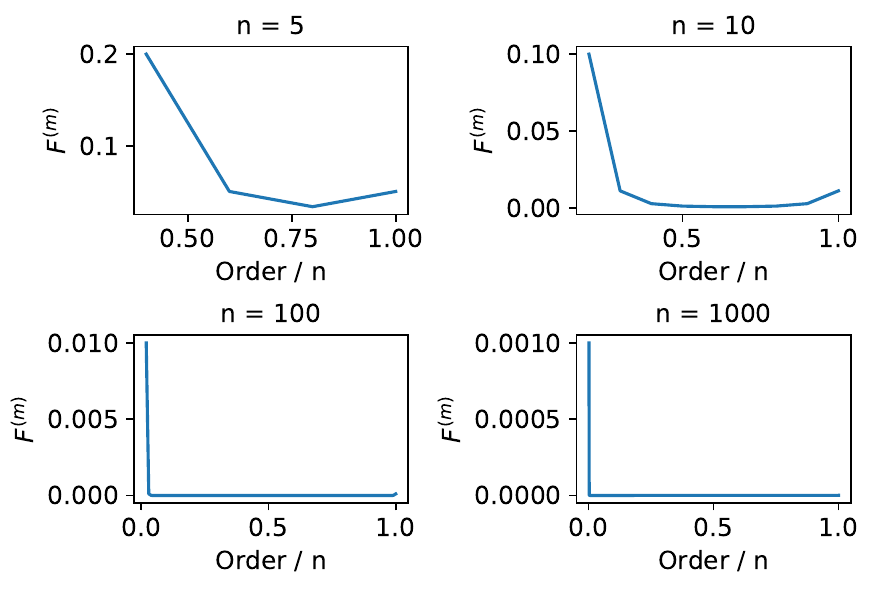}
\caption{Distributions of $F^{(m)}$ with different numbers of variables $n$ where $f(\emptyset)$ is taken into consideration.}
\label{fig: order_2}
\end{figure}

Fig.~\ref{fig: order_2} shows the theoretical distributions of $F^{(m)}$ for different $n$. Unlike Fig.~\ref{fig:str_order_cnn_mlp} (a), the empty set $\emptyset$ is allowed as input for DNNs. Apparently, when the number of variables $n$ is very large ($n\geq 100$), $F^{(m)}$ is only positive for $m/n\rightarrow 0$. For macromolecules such as proteins, the number of atoms is usually greater than 10,000. If the theorem in~\cite{deng2021discovering} that the strengths $\Delta W^{(m)}(i, j)$ of learning the $m$-order interaction is strictly proportional to $F^{(m)}$ holds, DNNs would be impossible to pay attention to any middle-order interactions, which is proven to be critical for modeling protein-protein interactions~\cite{liu2020deep,das2021classification}. 

\subsection{Examination of the Normal Distribution Hypothesis}
\label{app: normal_test}
We use \emph{scipy.stats.normaltest} in the Scipy package~\cite{virtanen2020scipy} to test the null hypothesis that $\frac{\partial \Delta f(i, j, S)}{\partial W}$ comes from a normal distribution, i.e, $\frac{\partial \Delta f(i, j, S)}{\partial W} \sim \mathcal{N}\left(0, \sigma^{2}\right)$. This test is based on D'Agostino and Pearson's examination and combines skew and kurtosis to produce an omnibus test of normality.  The $p$-values of well-trained EGNN and Molformer on the Hamiltonian dynamics dataset are 1.97147e-11 and 2.38755e-10, respectively. The $p$-values of randomly initialized EGNN and Molformer on the Hamiltonian dynamics dataset are 2.41749e-12 and 9.78953e-07, separately. Therefore, we are highly confident in rejecting the null hypothesis (\emph{e.g.}, $\alpha=0.01$) and insist that $\frac{\partial \Delta f(i, j, S)}{\partial W}$ depends on the data distributions of downstream tasks and the backbone model architectures.

\section{CNNs and MLP-Mixer on Visual Tasks}
\label{app: visual}
To investigate the change of interaction strengths during the training process in image classification, we train ResNet-50~\cite{he2016deep} and MLP-Mixer (\emph{Small})~\cite{nips2021mlpmixer} and calculate the interaction strength by the official implementation provided by~\cite{deng2021discovering}. MLP-mixer is an architecture based exclusively on multi-layer perceptrons (MLPs). It contains two types of layers: one with MLPs applied independently to image patches (\emph{i.e.} “mixing” the per-location features), and one with MLPs applied across patches (\emph{i.e.} “mixing” spatial information). We discuss the MLP-mixer to compare it with the traditional CNNs. Notably, CNNs assume the local inductive bias, while MLP-mixer instead connects each patch with other patches (\emph{e.g.}, no constraint of locality). 

We follow the training settings of DeiT~\cite{icml2021deit} and train 200 epochs with the input resolution of $224\times 224$ on CIFAR-10~\cite{krizhevsky2009learning} and ImageNet~\cite{russakovsky2015imagenet} datasets. 
Fig.~\ref{fig:str_order_cnn_mlp} plots the corresponding strengths at different epochs, where the dotted line denotes the initial interaction strength without training (referring to epoch 0), \emph{i.e.}, the data distribution of strengths $J^{(m)}_D$. 

Through visualization, it can be easily found that $J^{(m)}_D$ in both CIFAR-10 and ImageNet have already obeyed a mode that the low-order ($m/n\leq 0.2$) and high-order ($m/n\geq 0.8$) interaction strengths are much higher than middle-order ($0.2 \leq m/n\leq 0.8$). The variation of interaction strengths is very slight with the training proceeding, which validates our statement that the data distribution has a strong impact on the learned distribution. More importantly, we challenge the argument in~\cite{deng2021discovering}, who believe it is difficult for DNNs to encode middle-order interaction. However, in our experiments on GNNs, we document that DL-based models are capable of capturing middle-order interactions.

{The capability of CNNs to capture desirable levels of interactions is constrained by improper inductive bias. Remarkably, some preceding work~\cite{han2021demystifying,ding2022scaling} has proved the effectiveness of enlarging kernel size to resolve the local inductive bias, where adequately large kernel size can improve the performance of CNNs comparable to ViT and MLP-Mixer. Nevertheless, determining the scale of convolutional kernels is still under-explored. Our ISGR algorithm provides a promising way to seek the optimal kernel size based on the interaction strength, abandoning the exhaustive search.}  


\section*{Comprehensive Comparison to Existing Bottlenecks of GNNs}
\label{app: bottleneck_comparison}
    GNNs based on the message-passing diagram show extraordinary results with few layers. However, such GNNs fail to capture information that depends on the entire graph structure and prevent the information flow from reaching distant nodes. This phenomenon is called \textbf{under-reaching}~\cite{barcelo2020logical}. To overcome this limitation, an intuitive resolution is to increase the layers. However, unfortunately, GNNs with many layers tend to suffer from the \textbf{over-smoothing}~\cite{oono2019graph} or \textbf{over-squashing}~\cite{alon2020bottleneck} problems.
    
    \emph{Over-smoothing} takes place when node embeddings become indistinguishable. It occurs in GNNs that are used to tackle short-range tasks, \emph{i.e.}, the accurate prediction majorly depends on the local neighborhood. On the contrary, long-range tasks require as many layers as the range of interactions between nodes. But this would contribute to the exponential increase of the node's receptive field and compress the information flow, which is named \emph{over-squashing}. 
    Our study does not specify which problem category to be addressed (\emph{i.e.}, long-range or short-range), but explores which sort of interactions GNNs are more likely to encode (\emph{i.e.}, too simple, intermediately complex, and too complex). Note that different tasks require different levels of interaction. For instance, Newtonian and Hamiltonian dynamics demand too complex interactions, while molecular property prediction prefers interactions of intermediate complexity. Then, based on theoretical and empirical evidence, we discover that improper inductive bias introduced by the way to construct graph connectivity prevents GNNs from capturing the desirable interactions, resulting in the representation bottleneck of GNNs. 
    
    So what is the significant difference between our representation bottleneck and \emph{over-squashing}? Most fundamentally, our representation bottleneck is based on the theory of multi-order interactions, while \emph{over-squashing} relies on the message propagation of node representations. To be specific, it is demonstrated in~\cite{alon2020bottleneck} that the propagation of messages is controlled by a suitable power of $\hat{A}$ (the normalized augmented adjacency matrix), which relates \emph{over-squashing} to the graph topology. In contrast, we show that the graph topology strongly determines the distribution of interaction strengths $J^{(m)}$, \emph{i.e.}, whether GNNs are inclined to capture too simple or too complex interactions. This difference in theoretical basis leads to the following different behaviors of our representation bottleneck and \emph{over-squashing}:
    
    \begin{itemize}
        \item The multi-order interaction technique focuses on interactions under a certain context, whose complexity is measured as the number of its variables (\emph{i.e.}, nodes) $m$ divided by the total number of variables of the environment (\emph{i.e.}, the graph) $n$. Thus, the complexity of interactions is, indeed, a relative quantity. Conversely, \emph{over-squashing} (as well as \emph{under-reaching}) concerns about the absolute distance between nodes. Given a pair of nodes $i$ and $j$, if the shortest path between them is $r$, then at least $r$ layers are required for $i$ to reach out to $j$. More generally, long-range or short-range tasks discussed in most GNN studies refer to this $r$-distance. \emph{Over-squashing}, therefore, follows this $r$-distance metric and argues that the information aggregated across a long path is compressed, which causes the degradation of GNNs' performance. 
        
        As a result, our representation bottleneck can occur in both short-range and long-range tasks, but \emph{over-squashing} mainly exists in long-range problems. For short-range tasks, if we assume a KNN-graph with a large $K$ or even fully-connected graphs (\emph{i.e.}, nodes can have immediate interactions with distant nodes), then the receptive field of each node is very large, and GNNs intend to concentrate on too complex interactions but fail to capture interactions within local neighbors. For long-range tasks, if we assume a KNN-graph with a small $K$ (\emph{i.e.}, nodes only interact with nearby nodes), then the receptive field of each node is relatively small compared to the size of the entire graph. Consequently, GNNs prefer to capture interactions that are too simple but cannot seize more informative complex interactions. 
    
        \item More essentially, the multi-order interaction theory of our representation bottleneck is model-agnostic, but the starting point of \emph{over-squashing} is message passing, the characteristic of most GNN architectures. To make it more clear, the calculation of multi-order interactions (see Equ.~\ref{equ: multi_order_interaction} and~\ref{equ: strength}) is completely independent of the network (\emph{e.g.}, CNNs, GNNs, RNNs). However, the theory of \emph{over-squashing} is founded on the message-passing procedure. This hypothesis makes \emph{over-squashing} limited to the group of GNNs that are built on message passing. But other kinds of GNNs, such as the invariants of Transformers, may not suffer from this catastrophe. Instead, the analysis of multi-order interactions in our representation bottleneck can be utilized in any GNN architecture, even if it abandons the traditional message-passing mechanism. 
    \end{itemize}
    
    To summarize, our representation bottleneck is more universal than \emph{over-squashing}, which is built upon the absolute distance and merely talks about long-range tasks. This is because our representation bottleneck is given birth to by the theory of multi-order interactions rather than the property of message propagation. 
\end{document}